%% file: main.tex
\pdfoutput=1
\documentclass[10pt, a4paper, twocolumn]{article}

\usepackage[utf8]{inputenc}
\usepackage[T1]{fontenc}
\usepackage[english]{babel}

\usepackage{tgheros} 

\usepackage[varqu]{zi4}
\usepackage{microtype}

\usepackage{titlesec}
\titlespacing*{\section}{0pt}{1.5ex plus 1ex minus .2ex}{0.1em}
\titlespacing*{\subsection}{0pt}{1.5ex plus 1ex minus .2ex}{0.1em}
\titlespacing*{\subsubsection}{0pt}{1.5ex plus 1ex minus .2ex}{0.1em}
\titleformat*{\section}{\Large\bfseries\raggedright}

\usepackage[table]{xcolor}
\usepackage{booktabs}
\usepackage{float}
\usepackage{graphicx}
\usepackage{subcaption}
\usepackage[labelfont={bf,small}, textfont={small}]{caption}

\definecolor{acqua_celeste}{RGB}{0, 160, 175}

\usepackage{amsmath}
\usepackage{amssymb}
\usepackage{mathtools}
\usepackage{pifont}
\usepackage{multirow}
\usepackage{array}
\usepackage{makecell}
\usepackage{wrapfig}
\usepackage{tabularx}
\usepackage{xspace}

\usepackage[left=1.5cm, right=1.5cm, top=1.5cm, bottom=2cm]{geometry}
\usepackage[numbers, sort&compress]{natbib}
\usepackage{hyperref}
\usepackage[capitalize]{cleveref}
\crefname{section}{Sec.}{Secs.}
\Crefname{section}{Section}{Sections}
\Crefname{table}{Table}{Tables}
\crefname{appendix}{App.}{App.}\Crefname{appendix}{App.}{App.}\crefname{subappendix}{App.}{App.}\Crefname{subappendix}{App.}{App.}\crefname{subsubappendix}{App.}{App.}\Crefname{subsubappendix}{App.}{App.}\crefname{paragraph}{App.}{App.}\Crefname{paragraph}{App.}{App.}
\crefname{table}{Tab.}{Tabs.}

\hypersetup{
    colorlinks=true,
    linkcolor=black,
    urlcolor=acqua_celeste,
    citecolor=acqua_celeste
}

\usepackage{fancyhdr}
\usepackage{authblk}

\graphicspath{{figures/}{assets/proposals/}}
\newcommand{\measure}[1]{\textbf{\emph{#1}}}
\newcommand{\counterct}{CtrlCT}
\newcommand{\leadin}[1]{\par\textbf{#1}\enspace}
\makeatletter
\DeclareRobustCommand\onedot{\futurelet\@let@token\@onedot}
\def\@onedot{\ifx\@let@token.\else.\null\fi\xspace}
\def\eg{\emph{e.g}\onedot}
\def\ie{\emph{i.e}\onedot}
\def\vs{\emph{vs}\onedot}
\makeatother

\title{\huge\bfseries\vspace{-1em} One, No One and One Hundred Thousand CTs:\\Testing Anatomical Control in Text-to-CT Generation}

\author{Francesca Pia Panaccione}
\author{Eugenio Lomurno}
\author{Matteo Matteucci}

\affil{Politecnico di Milano, AIRLab, Italy}

\date{}

\begin{document}

\twocolumn[{%
  \begin{@twocolumnfalse}
    \maketitle
    \vspace{-3em}

    \begin{abstract}
        \setlength{\parindent}{0pt}
        \setlength{\parskip}{4pt}
        \itshape
        \noindent Recent advances in deep learning and the public availability of computed tomography (CT) datasets with paired reports have enabled many Text-to-CT generators.
        However, their evaluations almost always differ in preprocessing and metrics, which hinders comparison.
        Moreover, their protocols almost never test how editing the report changes the CT, not even at the same seed, so the report is often assumed to control the anatomy without evidence of how much and how consistently it does.
        We therefore propose \counterct{}, a toolkit that tests two properties of a generator, Credibility and Control.
        Credibility tests compare generated CTs, as a population, with real CTs and reports; Control tests are metamorphic and need no reference CT: at a fixed seed, editing one finding should change it as requested, keep negated findings absent and shift distant anatomy no more than a new seed, and the change should hold across seeds.
        A Report Card summarises all measures and insights for each generator.
        We benchmark five public generators on 940 chest CTs and 687 edits of pleural effusion and cardiomegaly, holding out a sixth to test the toolkit.
        Among them, the most credible generators barely respond to an added effusion, while the most responsive produce less realistic and less diverse CTs. The analyses proposed in the literature do not reveal this.
        We release \counterct{}, its edits, seeds and 30{,}525 synthetic CTs after acceptance.
    \end{abstract}

    \vspace{0.5em}

    \vspace{1em}
    \hrule height 1pt
    \vspace{1.5em}

    \centering
    \includegraphics[width=\textwidth]{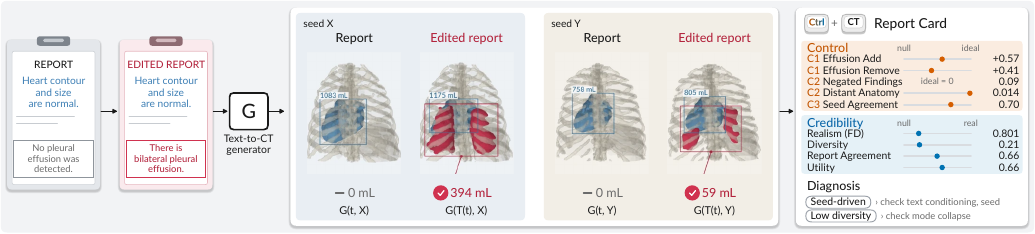}
    \captionof{figure}{\textbf{\counterct{} edits one finding of a real report and checks whether the CT follows.} The generator synthesises the CT of the report and of its edit at the same seed, here at two seeds (red: effusion; blue: heart; TotalSegmentator~\cite{Wasserthal_2023_TotalSegmentator}). Right: its \counterct{} Report Card over all cases and seeds (\cref{tab:results}): its main measures between their null control and real or ideal value, and the diagnostic rules that are triggered.}
    \label{fig:qual3d}
    \vspace{1.5em}
  \end{@twocolumnfalse}
}]

\section{Introduction}
\label{sec:introduction}

Text-to-CT generators synthesise 3D computed tomography (CT) volumes from radiology reports.
Advances in deep learning have led to a growing number of such generators~\cite{Molino_2025_Text2CT,Guo_2025_Text2CT_NVIDIA,Molino_2026_RAGText2CT,MedSyn2_2026}, most trained on CT-RATE~\cite{Hamamci_2024_CTRATE}, the reference public dataset of chest CTs paired with reports, released in 2024. Text-to-CT generation now also features in major medical imaging challenges, such as VLM3D~\cite{VLM3D_2025}, built on CT-RATE.
\providecommand{\Ty}{\makebox[1.15em][c]{\textcolor{tabok}{\ding{52}}}}
\providecommand{\Tn}{\makebox[1.15em][c]{\textcolor{tabno}{\ding{56}}}}
\providecommand{\Tp}{\makebox[1.9em][c]{\textcolor{tabpart}{(\ding{52})}}}
\providecommand{\Tnone}{\textcolor{black!35}{--}}
\providecommand{\Thd}[1]{\begin{tabular}[b]{@{}c@{}}#1\end{tabular}}
\providecommand{\Thl}[1]{\begin{tabular}[b]{@{}l@{}}#1\end{tabular}}
\definecolor{tabours}{HTML}{E6F2EF}\providecommand{\Tc}[2]{\multicolumn{1}{>{\columncolor{tabours}}#1}{#2}}
\definecolor{tabok}{HTML}{2E9E5B}\definecolor{tabno}{HTML}{D64545}\definecolor{tabpart}{HTML}{E3A21A}

\begin{table*}[t]
\centering
\footnotesize
\setlength{\tabcolsep}{4.2pt}
\renewcommand{\arraystretch}{1.1}
\caption{\textbf{Published evaluations of Text-to-CT generators versus \counterct{}.} \#Generators: other generators in the comparison tables of the paper (\counterct{}: five, plus CTFlow held out). \#Metrics: distinct metrics of those tables (lists in App.~A); a metric computed on several planes, windows or features, or reported as mean, median or standard deviation, counts once. Same-Seed Edits: CTs of a report and of its edit generated at the same seed and compared (in parentheses, the number of edits). Multiple Seeds: outcomes checked over several seeds. Real-CT Reference: metrics also computed on real CTs. Downstream Utility: synthetic CTs used to train a downstream model, tested on real CTs. Perturbed Reports: outputs tested with the report scrambled or replaced by unrelated text. \Ty\ yes; \Tp\ partly (App.~A); \Tn\ not reported.}
\label{tab:positioning}
\begin{tabular}{@{}l c c c c c c c@{}}
\toprule
Work & \#Generators & \#Metrics & Same-Seed Edits & Multiple Seeds & Real-CT Reference & Downstream Utility & Perturbed Reports \\
\midrule
GenerateCT~\cite{Hamamci_2024_GenerateCT} & 3 & 3 & \Tn & \Tn & \Tp & \Ty & \Tn \\
MedSyn~\cite{Xu_2024_MedSyn} & 4 & 3 & \Tn & \Tp & \Ty & \Ty & \Tn \\
Text2CT~\cite{Molino_2025_Text2CT} & 3 & 5 & \Tn & \Tp & \Ty & \Ty & \Tn \\
RAGText2CT~\cite{Molino_2026_RAGText2CT} & 5 & 5 & \Tn & \Tp & \Ty & \Tn & \Tn \\
MedSyn2~\cite{MedSyn2_2026} & 3 & 7 & \Tp & \Tp & \Ty & \Tp & \Tn \\
CTFlow~\cite{Wang_2025_CTFlow} & 4 & 3 & \Tn & \Tn & \Tn & \Tn & \Tn \\
\midrule
\multicolumn{1}{@{}>{\columncolor{tabours}[0pt][\tabcolsep]}l}{\textbf{\counterct{} (ours)}} & \Tc{c}{5+1} & \Tc{c}{12} & \Tc{c}{\Ty\,(687)} & \Tc{c}{\Ty} & \Tc{c}{\Ty} & \Tc{c}{\Ty} & \multicolumn{1}{>{\columncolor{tabours}[\tabcolsep][0pt]}c@{}}{\Ty} \\
\bottomrule
\end{tabular}
\end{table*}

Besides real reports, these generators can also be conditioned on modified versions of them, adapted to emulate the style and content of clinicians' reports and possibly describing rare cases. This is especially useful for rare combinations of findings, whether to augment real datasets without new annotated scans~\cite{Koetzier_2024_SyntheticData} or in clinical and simulation settings. A change in the report should produce the corresponding change in the generated CT, at every seed. Whether the report is real or edited, errors and misalignments between it and the CT can have a heavy impact, deciding whether these generators can be used in real cases.

Surprisingly, this correspondence is not systematically tested when new generators are released (\cref{tab:positioning}). Evaluations usually score sets of CTs from
unedited reports, by distance from real CTs (\eg, Fr\'echet distance~\cite{Heusel_2017_FID}) and agreement with the reports
(\eg, AUROC), not whether a CT changes when its report does.

Measuring the anatomical control of Text-to-CT generators, that is, quantifying and comparing their ability to modify a CT in a targeted way when its report changes, without altering its other regions, is currently an open challenge and a largely unexplored problem. The main difficulty is that an edited report describes a patient who was never scanned, so no reference CT exists. Without such tests, good scores can lead to an unjustified illusion of control over the generative process (\cref{sec:res-blind}). Beyond all this, a fair framework for comparing generators is missing, since each paper chooses its own baselines, preprocessing, features and scorers (App.~A).

We therefore build \counterct{}, a behavioural testing toolkit for Text-to-CT generators (\cref{fig:qual3d,fig:toolkit}), with two families of tests under one protocol. Pre-Flight Checks first harmonise the generated volumes: they align orientation and voxel spacing with real CTs and flag intensity offsets, missing CTs and edits beyond the text window (App.~D). Then, the \emph{Credibility tests} measure, on the CTs of unedited reports, how realistic and diverse they are against real CTs, how well a findings classifier recovers their reports, how much they move when the report is perturbed, and how useful they are to train classifiers tested on real CTs. Next, the \emph{Control tests} evaluate whether edits control the anatomy: they compare the CTs generated at a fixed seed from a report and from its edit, isolating the effect of the edit, and check three relations: the edited finding changes as requested (\emph{Target Response}, C1); findings that the report negates stay absent and distant anatomy shifts no more than under a new seed (\emph{Containment}, C2); and the response holds across seeds (\emph{Reliability}, C3). Finally, a Report Card summarises all measures and insights. The approach is designed to be reapplied to new generators.

We benchmark five public generators on chest CTs of CT-RATE, with edits that add or remove a \emph{pleural effusion} (fluid around the lungs) or \emph{cardiomegaly} (an enlarged heart), and use a sixth, CTFlow~\cite{Wang_2025_CTFlow}, held out to test the toolkit. Our results show that credible CTs can mask poor Control: the two most credible generators barely respond to an added effusion, despite using the report as a whole.

Our contributions are:
\begin{itemize}
  \item
  \textbf{A reference-free formalisation of anatomical control} through three metamorphic criteria, distinguishing targeted responses, unintended anatomical shifts and consistency across seeds.
  \item
  \textbf{Evidence that credible CTs can conceal failures of finding-specific control:} Realism and agreement with the whole report can be high when edits fail, as comparisons with conventional metrics on the same CTs show.
  \item
  \textbf{\counterct{}, a reusable toolkit for evaluating and diagnosing Text-to-CT generators} under a common protocol, demonstrated on five public models and independently applied to a sixth.
\end{itemize}

\section{Related Work}
\label{sec:related}
\leadin{Text-to-CT generators.}%
Text-to-CT generators differ in how they use report information for conditioning~\cite{Panaccione_2026_Survey,Panaccione_2026_OvESyn}: GenerateCT predicts CT tokens and then applies diffusion super-resolution~\cite{Hamamci_2024_GenerateCT}; CTFlow generates slice blocks by flow matching~\cite{Wang_2025_CTFlow}; Text2CT aligns reports and volumes contrastively~\cite{Molino_2025_Text2CT}; MedSyn and RAGText2CT add generated or retrieved anatomy~\cite{Xu_2024_MedSyn,Molino_2026_RAGText2CT}; MedSyn2 (MedSynV2) uses gated attention and optional masks~\cite{MedSyn2_2026}. Other generators, not evaluated here, lack public weights~\cite{Amirrajab_2025_Report2CT,Hamamci_2025_BTB3D,Guo_2025_Text2CT_NVIDIA} or are conditioned on masks with or without text~\cite{Guo_2025_MAISI} (we use MedSyn2 with text only; App.~A). Among the papers of \cref{tab:positioning}, two consider a change of report, without a systematic comparison at the same seed: MedSyn compares groups of CTs generated from a description and from its negation, on its own model~\cite{Xu_2024_MedSyn}, and MedSyn2 shows qualitative edits of report and masks, one at a fixed seed~\cite{MedSyn2_2026}. None compares the response to report edits across generators. 

\leadin{Benchmarks and evaluation protocols.}Benchmarks for synthetic chest radiographs~\cite{Dutt_2025_CheXGenBench} and for natural images and video~\cite{Huang_2024_CVPR,Liu_2024_CVPR,Pan_2025_ICCV} measure several properties under one fixed protocol, since scores depend on preprocessing and feature extraction~\cite{Parmar_2022_CVPR,Jayasumana_2024_CVPR,Woodland_2024_FeatureExtraction}, and release prompts and procedure instead of target outputs~\cite{Huang_2024_CVPR}. \counterct{} adapts this design to report-conditioned 3D CT, with report edits, report perturbations and multiple seeds.

\leadin{Controlled editing and behavioural testing.}Image editing benchmarks score whether the requested change appears and whether the rest of the input image is preserved, often within a mask~\cite{Ju_2024_PIEBench,Jia_2026_CompBench}. Prompt-to-Prompt edits generated images by changing the prompt and reusing cross-attention maps to preserve their layout~\cite{Hertz_2023_P2P}. Medical editing benchmarks~\cite{Liu_2025_MedEBench} and counterfactual studies~\cite{Gu_2023_BiomedJourney,Melistas_2024_Counterfactual,Monteiro_2023_Axiomatic,Yeganeh_2025_LatentDrifting} examine whether requested changes occur and what else changes. \counterct{} evaluates these behaviours in report-conditioned CT synthesis: it compares CTs generated from original and edited radiology reports at the same seed, checking whether the requested change occurs without unintended anatomical shifts, consistently across multiple seeds. Metamorphic testing formalises these checks as relations between inputs and outputs~\cite{Chen_2018_MetamorphicSurvey,Segura_2016_MTSurvey}, and CheckList~\cite{Ribeiro_2020_CheckList} adapted it to NLP models with directional tests, where an input change must move the output in a known direction, and invariance tests, where it must not; similar tests expose vision-language models that ignore how words are composed~\cite{Thrush_2022_CVPR,Yuksekgonul_2023_BagOfWords}. In \counterct{}, adding a finding is a directional test and rewording a report an invariance test (App.~C). 

\section{The \counterct{} Tests}
\label{sec:protocol}
\label{sec:setup}

\begin{figure}[t]
\centering
\includegraphics[width=0.6\linewidth]{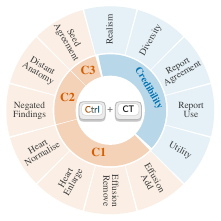}
\caption[The tests of CtrlCT]{\textbf{Criteria and measures of \counterct{}.} Inner ring: Credibility and the Control criteria C1--C3; outer ring: their measures.}
\label{fig:tests}
\end{figure}

\counterct{} evaluates two properties of a Text-to-CT generator (\cref{fig:tests}). \emph{Credibility}: as a population, the CTs generated from
real reports match the distribution of real CTs, agree with and depend on their reports, and are useful for training.
\emph{Control}: at a fixed seed, editing one finding changes it as requested and leaves the rest of the report
satisfied.

\label{sec:principles}
\leadin{Problem definition.}A generator $G_m$ is a frozen pipeline that maps a report $t$ and a seed
$s$ to a CT $x=G_m(t,s)$. For a fixed report, seeds draw CTs from a distribution $p_m(x\mid t)$.
A \emph{finding} $f$ is one of the 18 abnormalities labelled in CT-RATE~\cite{Hamamci_2024_CTRATE}. \emph{Evaluators} are models that assess findings in a CT: a classifier estimates finding probabilities, while a segmenter measures structure volumes where applicable (\cref{sec:evaluators}). Each numerical measurement is a \emph{readout} $h(x)$, while $c_f(x)$ denotes the state assigned to finding $f$ in $x$, such as present or absent. We edit a finding only if its evaluators perform reliably on real CTs and respond as expected to controlled image changes.

All generators receive the same real reports and seeds. Within each generator, we vary one input at a time: a statement, the whole report or the seed. Differences between the resulting CTs characterise the generator's response, without identifying the contributions of individual components of its pipeline.
Each measure is compared with up to two reference values: its value when the tested property is absent by design (\emph{null control}), and its value on real CTs or for an ideal generator.

\subsection{Credibility}
\label{sec:context}
Credibility is measured on the \emph{source CTs}, generated one per unedited report, and compared with real CTs. We use published metrics and add our own for behaviours that current approaches in the literature leave unexplored.
\begin{figure*}[t]
\centering
\includegraphics[width=\linewidth]{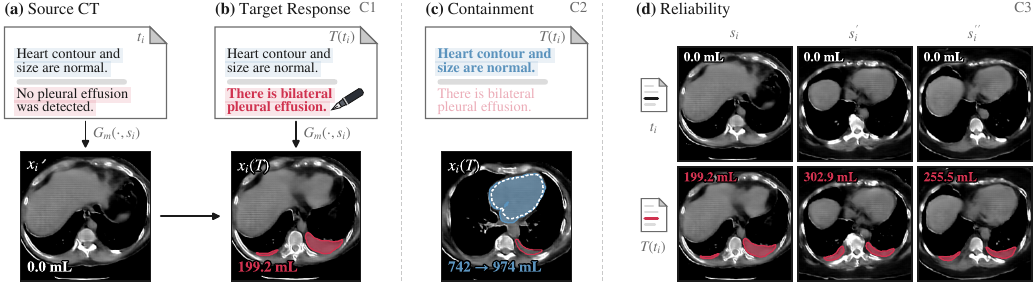}
\caption[Control on one case]{\textbf{Control on one case.} MedSyn generates $x_i$ from report $t_i$ and seed $s_i$ (a). Editing the effusion sentence at the same seed adds a bilateral effusion (b, C1); the heart, described as normal, grows (c, C2; dashed: before the edit). At two further seeds the original report yields no effusion and the edited report always yields one (d, C3). Red: effusion; blue: heart (TotalSegmentator~\cite{Wasserthal_2023_TotalSegmentator}). Case: the eligible addition closest to the median response of the five generators, by a rule fixed before viewing images (App.~A).}
\label{fig:walkthrough}
\end{figure*}

\measure{Realism} compares synthetic and real CTs as two populations, with the Fr\'echet distance (FD)~\cite{Heusel_2017_FID} on image features (\cref{sec:evaluators}). The FD is low between two volume-disjoint subsets of real data (real reference) and high when one of them is blurred with a Gaussian filter ($\sigma=4$\,mm), which makes the same CTs unrealistic (null control). The FD, however, also depends on \measure{Diversity}, how much of the real distribution the synthetic one covers. A generator can produce realistic CTs and still miss the rarer anatomies and details of the real data, the patterns that matter most when synthetic CTs are used in training sets. We separate the two on 3D features: precision~\cite{Kynkaanniemi_2019_PR}, the share of synthetic CTs within the neighbourhood of a real CT, measures Realism; recall~\cite{Kynkaanniemi_2019_PR} and coverage~\cite{Naeem_2020_PRDC}, the share of real CTs with a synthetic neighbour, measure Diversity. As for the FD, the reference of all three is their value between two disjoint subsets of real CTs; Diversity has no null control.

\measure{Report Agreement} checks that each generated CT shows the findings of its own report. Most published evaluations use a CLIP score or a findings classifier (App.~A); we compute the AUROC of a findings classifier against the report's labels for the 18 findings. Its null control is chance (0.5), and its reference is its value on real CTs. We then recompute \mbox{AUROC} after pairing each CT with the labels of another case's report (\emph{Swap}): the drop relative to the original \mbox{AUROC} quantifies how much better CTs agree with their own reports than with those of other cases. The drop is 0 when a CT agrees with any report as well as with its own (null control), and its reference is the drop on real CTs.

\measure{Report Use} measures how much the generated CTs change when their reports are perturbed. We introduce two modifications of the report: \emph{Scramble} permutes its words~\cite{Thrush_2022_CVPR} and \emph{Off-topic} replaces it with non-medical text. We compare the CTs generated from the perturbed reports with those generated from the real ones. To make the comparison as fair as possible, the generation seeds are fixed, so that only the reports vary. We define $r_P$ in normalised form, such that
\begin{equation}
  r_P=\frac{\sum_{i}d\bigl(x_i,\,G_m(P(t_i),s_i)\bigr)}{\sum_{i}d\bigl(x_i,\,x_{j(i)}\bigr)},
  \label{eq:report-use}
\end{equation}
where $x_i$ is the source CT of case $i$, $P$ the perturbation, $d$ the cosine distance in the same 3D features and $x_{j(i)}$ the source CT of another case. $r_P=0$ means that the perturbation produced no change in the generated volume, while $r_P=1$ that the perturbation moves the CT as far as another case's CT (scale checked on real CTs, App.~A). Neither end is desirable, and we take values near 0.5 as ideal (App.~C). A large $r_P$ means that the CT moves with the text, not that it follows it.

\measure{Utility} aims to measure how useful the generated data are for downstream tasks, and is computed following the train-on-synthetic, test-on-real (TSTR) paradigm~\cite{Dutt_2025_CheXGenBench}: for each finding, a binary logistic classifier is trained on the same 3D features of the generated CTs, labelled by the presence or absence of the finding in their reports, and tested on a held-out set of real CTs. The metric is the AUROC, and the null control corresponds to random guessing. The same classifier trained on features of real CTs is the reference.

\subsection{Control}
\label{sec:control}

No reference CT exists for an edited report, so we compare against each other the CTs that the generator produces from a report and from its edit. 
An \emph{edit} $T$ of a finding $f$ rewrites every statement about $f$ in a report $t_i$ and keeps the rest: it adds or removes
$f$, or changes a property, \eg a normal heart that becomes enlarged. $c^\star_T$ denotes the state of $f$ that $T$ requests, \eg present after an addition and absent after a removal. For each
case $i\in\mathcal I$ we use $K$ seeds. At each seed $s_{ik}$ we generate a CT from the report and one from its edit, $x_{ik}=G_m(t_i,s_{ik})$ and
$x_{ik}(T)=G_m(T(t_i),s_{ik})$, using common random numbers~\cite{Glasserman_1992_CRN}. The mean effect of the edit on a readout $h$ is
\begin{equation}
  \hat\Delta_h(T)=\frac{1}{|\mathcal I|K}\sum_{i\in\mathcal I}\sum_{k=1}^{K}\bigl[h(x_{ik}(T))-h(x_{ik})\bigr].
  \label{eq:estimator}
\end{equation}
\Cref{fig:walkthrough} shows the three metamorphic Control criteria, \mbox{C1--C3}, that instantiate this paradigm.

\leadin{C1 Target Response.}We measure whether a finding edited in the report changes accordingly in the generated CT (\cref{fig:walkthrough}b). We test the addition or removal of a pleural effusion and the enlargement or normalisation of the heart. An effusion counts as present from 5\,mL over both lungs, validated on real CTs (\cref{sec:evaluators}; same ranking at 10\,mL, App.~C). \measure{Effusion Add} and \measure{Effusion Remove} measure, respectively, the change after the edit in the share of CTs with an effusion and in the share of CTs without one. This is the \emph{net effect} $\hat\Delta_h(T)$ of \cref{eq:estimator} with the readout $h(x)=\mathbf{1}[c_f(x)=c^\star_T]$, which marks whether the CT is in the requested state. Both measures are 0 when the edit has no effect (null control) and 1 when every edited CT, and no original one, is in the requested state (ideal).

Since no validated threshold separates a normal from an enlarged heart, \measure{Heart Enlarge} and \measure{Heart Normalise} instead measure the mean change of the heart volume $V$ after the edit, on a log scale: $\hat\Delta_h(T)$ with $h=\log V$ for an enlargement and $h=-\log V$ for a normalisation, so that a positive value indicates a change in the requested direction. Their null control is again 0, and their reference is computed from real CTs ($\Delta\log V=0.37$, App.~A).

\begin{figure}[t]
\centering
\includegraphics[width=\linewidth]{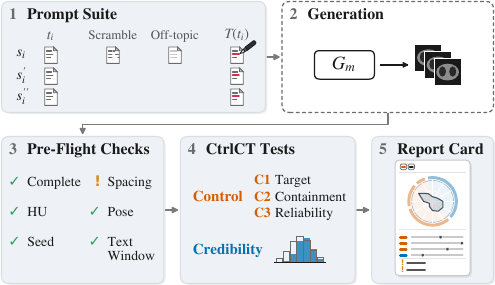}
\caption[The CtrlCT toolkit]{\textbf{The \counterct{} toolkit} (\cref{sec:toolkit-setting}). Five steps: the Prompt Suite is passed to an external generator, its CTs are checked, the tests are computed and the results are summarised in the Report Card.}
\label{fig:toolkit}
\end{figure}

\leadin{C2 Containment.}We measure whether an edit changes the generated CT only where requested (\cref{fig:walkthrough}c). We test two requirements: negated findings must stay absent, and anatomy far from the edited finding must not shift more than it varies between CTs of the unedited report generated with different seeds. 
\measure{Negated Findings} tests the first requirement. For each finding negated in the report, a classifier checks its presence in the CT of the report and in that of its edit, both generated with the same seed. The measure is how much their presence rises after the edit, on average (\cref{eq:estimator}). We remove the part of this rise due to the classifier, which tends to detect the findings that usually accompany an effusion (\cref{sec:evaluators}). Its ideal value is 0, \ie no negated finding appears; it has no null control. We do not count findings that the report does not mention: in real patients they can appear together with $f$, so they may legitimately change with it (\cref{sec:res-specificity}).
\measure{Distant Anatomy} tests the second: an effusion legitimately displaces the adjacent tissue, so we read the anatomy beyond
50\,mm from it (App.~A), where the edit requests no change.
Paired MAE and SSIM~\cite{Ju_2024_PIEBench} compare the two CTs of a single seed, so they cannot tell a change caused by the edit from one that a new seed would also cause. We therefore compare the effect of the edit with that of the seed. For the voxel intensities $g$ of this region, with mean and total variance over seeds
$\mu^{(0)}_{g,i},v^{(0)}_{g,i}$ under $t_i$ and $\mu^{(1)}_{g,i},v^{(1)}_{g,i}$ under $T(t_i)$, we use Fisher's discriminant
ratio~\cite{Fisher_1936_Discriminant}, pooled over cases:
\begin{equation}
  \theta_g=\frac{\sum_{i\in\mathcal I}\|\mu_{g,i}^{(1)}-\mu_{g,i}^{(0)}\|_2^2}
  {\sum_{i\in\mathcal I}\bigl(v_{g,i}^{(0)}+v_{g,i}^{(1)}\bigr)}.
  \label{eq:theta}
\end{equation}
Values $\theta_g>1$ mean that the edit shifts the mean of $g$ more than the seed makes $g$ vary. Its ideal is 0, no systematic shift; $\theta_g$ has no null control. We report $\theta_+$ for additions and $\theta_-$ for removals. A small $\theta_g$ means that the edit changes the distant anatomy no more than a new seed does, even if the two CTs differ. With only $K$ seeds per case, we remove the seed noise from the estimated shift and resample patients for the intervals (App.~A).

\leadin{C3 Reliability.}Whether an edit succeeds should depend on the case, not on the seed, although shape and volume may vary (\cref{fig:walkthrough}d). \measure{Seed Agreement} is Fleiss' $\kappa$~\cite{Fleiss_1971_Kappa} of the edit's success over the $K=3$ seeds, on the cases eligible at every seed (for an addition, all CTs of the original report lack $f$): $\kappa=1$ (ideal) if the seeds agree on every case, $\kappa=0$ (null control) if they agree only as often as chance, given the success rate. Consistent failures also count as agreement, so $\kappa$ is read with C1. \Cref{tab:results} reports effusion additions ($\kappa_+$; 253--356 eligible cases per generator; removals and heart edits in App.~C).

\section{Experimental Setting}
\label{sec:validation}
\label{sec:setting}
\label{sec:toolkit-method}

\subsection{Cohort and Edits}
\label{sec:cohort}
CT-RATE~\cite{Hamamci_2024_CTRATE} pairs non-contrast chest CTs with reports and 18 abnormality labels extracted from them. From
the validation reconstructions listed by the Text2CT release~\cite{Molino_2025_Text2CT}, we keep one per scan: 940 CTs of 782
patients, none of whom appears in the training split.
We edit two findings and fix all edited texts before generating any CT. For pleural effusion we use the 567 reports (479 patients) whose statement on effusion can be replaced without touching another finding: 396 negate it, and the edit adds it; 171 state it, and the edit removes it. For heart size we use 120 reports, one per
patient: 50 enlargements and 70 normalisations. An addition requests a combination of findings that is rare in the CT-RATE training split (24{,}128 scans). For the 396 effusion additions, the median number of training scans with the same 18 labels falls from 10.5 for the original report to 1 for the edited one, and no training scan matches the edited report in 41\% of them (62\% of heart enlargements). Rare is not unrealistic: 30\% of the 940 unedited reports also match no training scan. \Cref{fig:qual3d} shows one addition; App.~A provides further details.

\begin{table*}[t]
\centering
\captionsetup{skip=3pt}
\captionof{table}{\textbf{Results of \counterct{}} (left), generators in order of publication; no overall score. C1: net effect on effusion presence; change of log heart volume. C2: rise in calls of negated findings after an addition; $\theta$ beyond 50\,mm after an addition ($+$) or removal ($-$). C3: Fleiss' $\kappa$ of an addition's success over three seeds. Grey: null control, and value on real CTs (heart edits; Credibility except Report Use) or ideal (other rows); --: no null control. Bold: best; underlined: second (intervals: App.~C). Read C2 and C3 with C1: a generator that ignores the edit changes nothing and fails at every seed. FD: 200 synthetic vs 692 real CTs. Rows: measures of \cref{fig:tests}, metric in brackets; variants indented; for Report Use (0: text ignored; 1: as far as another case's CT), 0 and $\geq$1 are both failures and best is closest to 0.5 (App.~C).}
\label{tab:results}
\sbox0{\footnotesize\setlength{\tabcolsep}{1.7pt}\renewcommand{\arraystretch}{0.90}\begin{tabular}{@{}lccccccc@{}}
\toprule
Measure & {\scriptsize \textcolor{gray}{\textit{Null}}} & {\scriptsize \textcolor{gray}{\textit{Real or Ideal}}} & {\scriptsize GenerateCT} & {\scriptsize MedSyn} & {\scriptsize Text2CT} & {\scriptsize MedSyn2} & {\scriptsize RAGText2CT} \\
\midrule
\multicolumn{8}{@{}l}{{\setlength{\fboxsep}{1.1pt}\colorbox[HTML]{F6DCC7}{\textcolor[HTML]{D55E00}{\textbf{C1}}}}~\textcolor[HTML]{D55E00}{Target Response}} \\
Effusion Add$\uparrow$ & \textcolor{gray}{0.00} & \textcolor{gray}{1.00} & +0.20 & \textbf{+0.57} & +0.02 & \underline{+0.47} & +0.05 \\
Effusion Remove$\uparrow$ & \textcolor{gray}{0.00} & \textcolor{gray}{1.00} & +0.09 & \textbf{+0.41} & +0.21 & \underline{+0.27} & +0.18 \\
Heart Enlarge$\uparrow$ & \textcolor{gray}{0.00} & \textcolor{gray}{+0.37} & $-$0.02 & \underline{+0.15} & \textbf{+0.16} & +0.07 & +0.13 \\
Heart Normalise$\uparrow$ & \textcolor{gray}{0.00} & \textcolor{gray}{+0.37} & +0.00 & +0.13 & \textbf{+0.21} & +0.06 & \underline{+0.20} \\
\addlinespace[1.5pt]
\multicolumn{8}{@{}l}{{\setlength{\fboxsep}{1.1pt}\colorbox[HTML]{F6DCC7}{\textcolor[HTML]{D55E00}{\textbf{C2}}}}~\textcolor[HTML]{D55E00}{Containment}} \\
Negated Findings$\downarrow$ & \textcolor{gray}{--} & \textcolor{gray}{0.00} & \underline{0.02} & 0.09 & 0.03 & 0.04 & \textbf{0.01} \\
Distant Anatomy ($\theta_+$)$\downarrow$ & \textcolor{gray}{--} & \textcolor{gray}{0.000} & 2.87 & \underline{0.014} & 0.076 & \textbf{0.005} & 1.79 \\
\quad ($\theta_-$)$\downarrow$ & \textcolor{gray}{--} & \textcolor{gray}{0.000} & 3.20 & \underline{0.029} & 0.10 & \textbf{0.003} & 1.59 \\
\addlinespace[1.5pt]
\multicolumn{8}{@{}l}{{\setlength{\fboxsep}{1.1pt}\colorbox[HTML]{F6DCC7}{\textcolor[HTML]{D55E00}{\textbf{C3}}}}~\textcolor[HTML]{D55E00}{Reliability}} \\
Seed Agreement ($\kappa_+$)$\uparrow$ & \textcolor{gray}{0.00} & \textcolor{gray}{1.00} & \textbf{0.70} & \textbf{0.70} & 0.10 & 0.36 & 0.62 \\
\addlinespace[1.5pt]
\multicolumn{8}{@{}l}{{\setlength{\fboxsep}{1.1pt}\colorbox[HTML]{C7E0EE}{\textcolor[HTML]{0072B2}{\textbf{Credibility}}}}} \\
Realism (FD)$\downarrow$ & \textcolor{gray}{1.910} & \textcolor{gray}{0.035} & 1.412 & 0.801 & \underline{0.632} & 0.726 & \textbf{0.531} \\
Diversity (coverage)$\uparrow$ & \textcolor{gray}{--} & \textcolor{gray}{0.94} & 0.07 & 0.21 & \underline{0.43} & 0.27 & \textbf{0.44} \\
Report Agreement (AUROC)$\uparrow$ & \textcolor{gray}{0.50} & \textcolor{gray}{0.85} & 0.52 & 0.66 & \textbf{0.74} & 0.62 & \underline{0.73} \\
\quad (drop under Swap)$\uparrow$ & \textcolor{gray}{0.00} & \textcolor{gray}{0.35} & 0.04 & 0.17 & \textbf{0.24} & 0.13 & \underline{0.22} \\
Report Use ($r_P$, Off-topic) & \textcolor{gray}{0/$\geq$1} & \textcolor{gray}{$\approx$0.5} & 1.02 & \underline{0.68} & 0.93 & \textbf{0.38} & 0.94 \\
\quad ($r_P$, Scramble) & \textcolor{gray}{0/$\geq$1} & \textcolor{gray}{$\approx$0.5} & 1.08 & \underline{0.60} & \textbf{0.52} & 0.36 & 0.65 \\
Utility (TSTR AUROC)$\uparrow$ & \textcolor{gray}{0.50} & \textcolor{gray}{0.78} & 0.54 & 0.66 & \textbf{0.69} & 0.66 & \underline{0.67} \\
\bottomrule
\end{tabular}}%
\begin{minipage}[c]{\wd0}\usebox0\end{minipage}\hfill
\begin{minipage}[c]{\dimexpr\textwidth-\wd0-6pt\relax}\raggedleft
\includegraphics[height=\dimexpr\ht0+\dp0\relax,width=\linewidth,keepaspectratio]{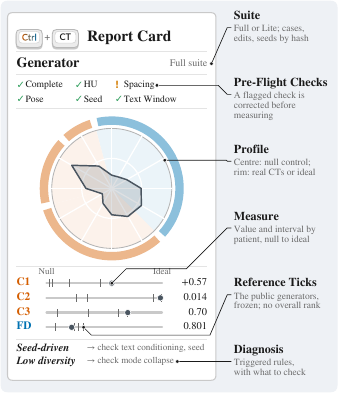}
\end{minipage}
\captionsetup{skip=2pt}
\captionof{figure}{\textbf{The \counterct{} Report Card} (right) of one of the five generators (illustrative profile).}
\label{fig:card}
\end{table*}

\subsection{Generators}
\label{sec:generators}
We benchmark five of the six public generators of \cref{tab:positioning} with their official code and weights; where code and
paper disagree, we follow the code. The sixth, CTFlow, is held out to test the toolkit (\cref{sec:toolkit}). Generators differ in orientation: we read each generator's orientation from the positions of heart, spine, lungs and liver segmented by
TotalSegmentator~\cite{Wasserthal_2023_TotalSegmentator}, and flip its CTs to match real CTs. For each edit, the original and the edited report are generated at the same $K=3$ seeds of the case, $s_{ik}$ (\cref{sec:control}), obtained by SHA-256 from the case identifier and $k$ (App.~A); Credibility measures use only $s_{i1}$, the $s_i$ of \cref{sec:context}. Seeds are fixed before generation and shared by all generators, which together produce 30{,}525 CTs.

All generators receive the same full report, although they were trained on different texts (App.~A); text encoders truncate it to a window of 256 tokens (GenerateCT), 512 (Text2CT,
RAGText2CT) or 2{,}048 (MedSyn, MedSyn2). Of the 567 effusion edits, 148 fall beyond the window for GenerateCT, one each for
Text2CT and RAGText2CT, none for MedSyn and MedSyn2. The text encoder drops such an edit, so the CT of the edit is voxel-identical to that of the original report;
we count it as a failure of the released pipeline. App.~C reports results within the window.

\subsection{Evaluators}
\label{sec:evaluators}
\leadin{Feature evaluators.}A feature evaluator maps a whole CT to features. The FD of Realism uses 2.5D RadImageNet features~\cite{Mei_2022_RadImageNet}, as in Text2CT and MedSyn2; all other feature-based measures (Diversity, Report Use, Utility) use the 3D \mbox{Pillar-0} features~\cite{Agrawal_2025_Pillar0} adopted by MedSyn2. Features have no labels, so we check them on real CTs: two disjoint groups of real CTs have a low FD
(0.035), and Pillar-0 reproduces its published value (App.~B).

\leadin{Finding evaluators.}For Report Agreement we use
NV-Reason-CT~\cite{Myronenko_2026_NVReasonCT}, the best-performing public classifier of the 18 CT-RATE findings that we tested (macro AUROC 0.85 on real CTs, against 0.68 for CT-CLIP~\cite{Hamamci_2024_CTRATE}). A classifier can also respond to signs that usually accompany a
finding: on real CTs, NV-Reason-CT's calls of cardiomegaly rise with pleural effusion by 0.14 more than the report labels do (its excess), so
after an added effusion it may call a cardiomegaly that the CT does not show. For Control we therefore measure the edited structure, effusion or heart volume in mL, with TotalSegmentator~\cite{Wasserthal_2023_TotalSegmentator} (Effusion Add/Remove, Heart Enlarge/Normalise, Seed Agreement and the region of Distant Anatomy), and keep the classifier only for Negated Findings, net of its excess. We edit only findings that both evaluators detect reliably on real CTs against the report labels (AUROC $\geq0.80$ or Youden's $J\geq0.50$): pleural effusion (volume: $J=0.86$) and cardiomegaly (heart volume: AUROC 0.83), not lung nodules.

\leadin{Evaluator checks.}Each evaluator is checked on real CTs, as above, and under known image changes (blur, mirroring, an HU offset), to which it responds as expected (App.~B). On synthetic CTs, without expert annotation, the two finding evaluators agree on both edited findings: Cohen's $\kappa$ 0.66--0.88 on effusion (0.89 on real CTs), AUROC 0.84--0.93 between heart volume and cardiomegaly calls, except GenerateCT (0.54--0.58, 0.60).

\subsection{\counterct{} Toolkit}
\label{sec:toolkit-setting}
\begin{figure*}[t]
\centering
\includegraphics[width=\linewidth]{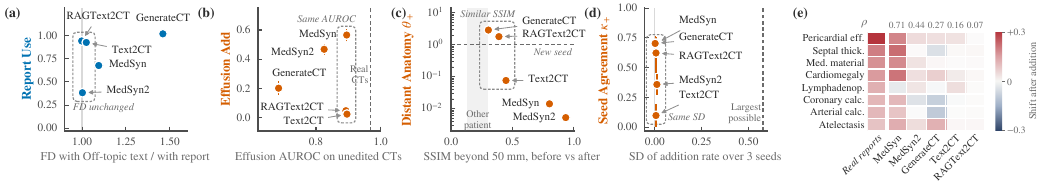}
\caption{\textbf{Published metrics and \counterct{} measures.} (a)--(d)~Each point is a generator: horizontally, a published metric; vertically, the corresponding \counterct{} measure (colours as in \cref{fig:tests}; bars: 95\% bootstrap intervals over patients). Dashed boxes: generators that the published metric rates alike. (e)~After an added effusion, shift in the classifier's call rate of each of the eight findings that co-occur most with effusion in real reports, net of its excess (pale: 95\% interval includes zero). Real reports: co-occurrence with effusion (log odds ratio $\times\,0.12$). $\rho$: Spearman correlation of each generator's shifts with this co-occurrence, over all 17 other findings.}
\label{fig:standard}
\end{figure*}

The \counterct{} toolkit is implemented in five steps (\cref{fig:toolkit}). (1)~During the first step, the Prompt Suite is produced, \ie the set of reports and seeds to be given as input to the generative process. (2)~The generator uses this information to produce the synthetic volumes. (3)~The Pre-Flight Checks align the orientation of these volumes and flag what would invalidate the measures. (4)~The \counterct{} Tests compute each measure with its 95\% bootstrap interval over patients. (5)~The Report Card (\cref{fig:card}) is the output of the toolkit: it places each measure between its null control and its real or ideal value, and lists the diagnostic rules that are triggered. The twelve rules are calibrated on the five benchmarked generators and applied to the held-out CTFlow as a user would to a new generator. App.~D gives the details and the cards of all the 5+1 generators.

The Prompt Suite has two versions: Full, with 6{,}017 CTs per generator (the 940 cases, 567 effusion edits and 120 heart edits), used for the benchmark, and Lite, with 1{,}814 (200 cases, 100 effusion edits and the same 120 heart edits), for cheaper tests of new generators. Lite is a random subset of the cases and effusion edits: over 50 draws of it, the rankings of the five generators agree with those of Full on the measures compared (mean Kendall's $\tau$ 0.78--1.00; App.~D). 

\section{Results}
\label{sec:results}

\label{sec:res-main}
\label{sec:res-edit}

\leadin{No generator reaches real CTs on Credibility.}\Cref{tab:results} shows the overall performance of the five benchmarked generators. In its Credibility rows, the FD is 15 to 40 times its value on real CTs, coverage is at most 0.44 (0.94 on real CTs), Report Agreement drops by only 0.04--0.24 under Swap (0.35 on real CTs), and classifiers trained on synthetic CTs obtain an AUROC of 0.54--0.69 when tested on real CTs, against 0.78 when the training set is real. RAGText2CT and Text2CT are the top two on Realism, Diversity and Report Agreement.

\leadin{Credibility and Control dissociate.}RAGText2CT and Text2CT have the lowest Effusion Add: when the effusion sentence is added, the proportion of CTs with an effusion rises from 7\% to 12\% and from 7\% to 9\%, compared with 5\% to 62\% for MedSyn, fourth on Realism. RAGText2CT and Text2CT, instead, remove an effusion more often than they add one (Effusion Remove $+0.18$ and $+0.21$); GenerateCT does the opposite (Effusion Add $+0.20$, Effusion Remove $+0.09$).

\leadin{Control depends on the finding.}In the C1 rows of \cref{tab:results}, Text2CT, last on Effusion Add ($+0.02$), is among the most responsive generators on the heart (Heart Enlarge $+0.16$, Heart Normalise $+0.21$). MedSyn2, second on both effusion measures, is fourth on both heart measures ($+0.07$, $+0.06$), while the mean volume of the hearts generated by GenerateCT does not change detectably with either edit ($-0.02$, $+0.00$). Control therefore depends on the edited finding: a ranking obtained by editing one finding does not carry over to another. None reaches the ideal or real value on C1: at best, Effusion Add is $+0.57$ (ideal 1) and a heart edit 57\% of the value on real CTs ($+0.21$ vs.\ $+0.37$).

\leadin{No generator is best on every Control measure.}MedSyn, best on Effusion Add, is worst on Negated Findings (0.09 [0.07, 0.11]). MedSyn2 is best on Distant Anatomy ($\theta_+=0.005$) but fourth on Seed Agreement ($\kappa_+=0.36$): whether an added effusion appears often depends on the seed. GenerateCT ties with MedSyn on Seed Agreement (\mbox{$\kappa_+=0.70$}; 0.65 on the edits it reads) but is worst on Distant Anatomy ($\theta_+=2.87$). RAGText2CT rarely adds the effusion ($+0.05$), yet its edits shift the distant anatomy more than a new seed would ($\theta_+=1.79$; App.~C).

\leadin{The held-out evaluation.}\label{sec:toolkit}We run CTFlow~\cite{Wang_2025_CTFlow}, the held-out generator on which the rules were not calibrated, on the Lite suite. Again, Realism and Control dissociate: CTFlow has the lowest FD (0.31), but ranks third of six on Effusion Add ($+0.24$; $+0.44$ on the edits it reads: 52 of 100 fall beyond its 128-token window), and the report explains 13\% of its feature variance over cases and seeds (13--64\% for the five generators; App.~A). The Pre-Flight Checks flag \emph{Edits not read}: the 52 edits beyond its window give the same CT as the unedited report. The Report Card adds \emph{Low diversity} (coverage 0.25) and, as borderline, \emph{Seed-driven}, since a new seed changes the CT as much as Off-topic text, and \emph{Rewrites the chest when editing}, since the distant anatomy changes almost as much as for another patient (App.~D).

\subsection{Comparison with Published Metrics}
\label{sec:res-blind}

Published metrics do not capture the discrepancies revealed by \counterct{}. \Cref{fig:standard}a--d shows, on the same CTs, four published metrics against the corresponding measures computed by the toolkit: FD, the AUROC of a findings classifier, a population measure repeated over seeds and paired SSIM~\cite{Ju_2024_PIEBench,Jia_2026_CompBench}.

\leadin{FD misses report dependence (\cref{fig:standard}a).}Replacing the report with Off-topic text changes the FD by less than 3\% for RAGText2CT, Text2CT and MedSyn2. Report Use, which compares each CT with the one generated from Off-topic text for the same case and seed, separates them: the CTs of the first two change almost as much as with another case's report (0.94 and 0.93), those of MedSyn2 much less (0.38). FD can compare populations as a whole, but gives no information on their content, which can vary widely for the same FD.

\leadin{Findings AUROC misses the response to an edit (\cref{fig:standard}b).}On the source CTs, a classifier's effusion AUROC against the report labels rates Text2CT and MedSyn alike (0.90, 0.89): both draw an effusion for the patients whose reports describe one. But those reports differ in many other findings. When only the effusion sentence is added, at the same seed, an effusion appears with MedSyn and hardly ever with Text2CT (Effusion Add $+0.57$ and $+0.02$). The published metric does not register this even on the edited reports: scored against them, the macro AUROC over the 18 findings differs by less than 0.005 between the edited CTs and the original ones, which ignore the edit; only the AUROC of the edited finding differs: 0.70 for MedSyn and 0.18 for Text2CT, below chance because most of its edited CTs keep the effusion state of the original report.

\leadin{Paired SSIM misses systematic shifts (\cref{fig:standard}c).}SSIM beyond 50\,mm from the effusion compares the CTs before and after the edit at the same seed. It rates Text2CT close to RAGText2CT and GenerateCT (0.45, 0.39, 0.30): in all three the distant anatomy changes. But a low SSIM does not tell whether the edit moves this anatomy the same way at every seed, a systematic shift, or only changes it as much as a new seed would. Distant Anatomy can: $\theta_+$ is 0.076 for Text2CT, whose shift stays within the variation across seeds, and 1.79 and 2.87 for RAGText2CT and GenerateCT, whose shift exceeds it.

\leadin{Rates over seeds miss seed-decided cases (\cref{fig:standard}d).}Repeated over the three seeds, the proportion of successful additions has a standard deviation of at most 0.011: as a population, every generator is stable. Single additions are not. With MedSyn, 22\% of eligible additions succeed at some seeds and fail at others, and with MedSyn2 48\%, while Text2CT and RAGText2CT are consistent only in failing (91\% of their eligible additions fail at every seed). Seed Agreement separates MedSyn from MedSyn2 ($\kappa_+$ 0.70 and 0.36).

\section{Insights}
\label{sec:res-insights}

\leadin{One sentence does not control one finding.}\label{sec:res-specificity}Whether a finding appears depends on the other findings of the report, not only on its own sentence. When the edit fails, the rest of the report decides: among the removals, which all use the sentence ``No pleural effusion was detected.'', an effusion remains in 45\% of Text2CT's CTs when the other findings make one unlikely (lowest third) and in 74\% when they make one likely (highest third; App.~C). When the edit succeeds, other findings change with it: for MedSyn and MedSyn2, 14 and 13 of the 17 other findings shift after an added effusion as they co-occur with effusion in real reports (Spearman 0.71 and 0.44; \cref{fig:standard}e), including findings that the report negates. Both results are consistent with generators that learn associations between whole reports and whole CTs.

\leadin{Text content weakly controls the CT.}\label{sec:res-noise}With the report replaced by Off-topic text, the CTs of GenerateCT, RAGText2CT and Text2CT change about as much as with another case's report (Report Use 1.02, 0.94 and 0.93), and still by 1.08, 0.65 and 0.52 when its words are only permuted (Scramble); these three have the lowest Effusion Add ($+0.20$, $+0.05$ and $+0.02$). MedSyn and MedSyn2 change less (0.68 and 0.38) and have the highest Effusion Add ($+0.57$ and $+0.47$). A strong response to the text is therefore no sign of Control: text describing no chest changes their CTs only as much as another case's report. The change stays limited even for a heavily corrupted text, which lets us state that their knowledge is weakly controlled by the text content.

\leadin{Implications for evaluation.}Realism, Report Agreement and Utility can be high even while an edit fails. At the current state of the art, a CT generated from an edited report cannot be assumed to match its label. Edits should be granularly evaluated: per finding, at the same seed and across seeds, against the findings that the report negates, and with the text window reported.

\section{Limitations, Outlook and Conclusion}
\label{sec:discussion}

\leadin{Limitations.}Our evaluators are automatic: no radiologist read the synthetic CTs. The study is vertical on chest CT, with one dataset and no other anatomy or modality. We edit two findings, among those that both evaluators detect reliably. Edits use three seeds, and the rules are calibrated on five generators and tested on one held-out generator.

\leadin{Future directions and conclusion.}We presented \counterct{}, a unified framework for the fair evaluation of Text-to-CT generators, and found serious gaps in current evaluations: the most credible generators are not the most controllable, and published metrics miss it. Next steps are other domains and, above all, generators that account for all these dimensions, to be used with more confidence and control and to advance the field.

\section*{LLM Usage Statement}
The research question, the protocol, the experiments and the interpretation of the results are the authors' own. We used Claude Code (Anthropic) as an assistant to help improve the writing and, under the authors' guidance, to support coding. The authors verified all results, revised the manuscript and take full responsibility for its content.

\bibliographystyle{plainnat}
\bibliography{references}

\clearpage
\appendix
\setcounter{figure}{0}\setcounter{table}{0}\setcounter{equation}{0}
\renewcommand{\thefigure}{A\arabic{figure}}
\renewcommand{\thetable}{A\arabic{table}}
\renewcommand{\theequation}{A\arabic{equation}}
\setcounter{dbltopnumber}{3}\renewcommand{\dbltopfraction}{0.85}\renewcommand{\dblfloatpagefraction}{0.7}
\setcounter{topnumber}{3}\renewcommand{\topfraction}{0.85}\renewcommand{\floatpagefraction}{0.7}
\crefalias{section}{appendix}\crefalias{subsection}{subappendix}\crefalias{subsubsection}{subsubappendix}
\noindent This appendix provides additional protocol details and analyses supporting the main paper.
Appendix tables, figures, and equations are prefixed with A; unprefixed references refer to the main paper.

\input{appendix/e01_protocol}
\input{appendix/e02_evaluators}
\input{appendix/e03_results}

\input{appendix/e04_toolkit}

\end{document}

%% file: appendix/e01_protocol.tex
\section{Evaluations, Setting and Measures}
\label{sec:supp-protocol}

\subsection{Published Evaluations}
\label{sec:supp-published}

\begin{figure*}[t]
\centering
\includegraphics[width=0.86\linewidth]{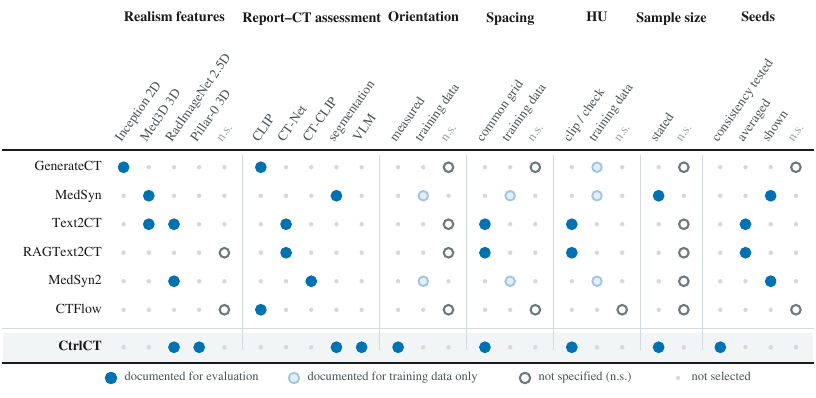}
\caption{\textbf{Evaluation protocols reported by previous Text-to-CT studies and CtrlCT.} Blue: setting documented for evaluation; pale blue: documented only for training data; open grey circle: not specified (n.s.); small grey dot: not selected. Spacing: common
grid, all CTs resampled to one voxel grid before evaluation.}
\label{fig:supp-choices}
\end{figure*}

\cref{tab:supp-published} provides further details on the generators and metrics presented in Tab.~1, specifying
which evaluation criteria are only partly met. \cref{fig:supp-choices} compares their evaluation protocols with that
of \counterct{}.

\begin{table*}[t]
\centering\footnotesize\setlength{\tabcolsep}{3.5pt}
\caption{\textbf{Details of Tab.~1.} Baselines and metrics of each study's comparison tables, counted as in Tab.~1,
and why a criterion is only partly met. Census closed on 22 October 2026.}
\label{tab:supp-published}
\input{assets/proposals/supplement/tables/published.tex}
\end{table*}

\subsection{Cohort and Edits}
\label{sec:supp-cohort}

\cref{tab:supp-sets} details the cohort, the edits and the perturbations. \cref{tab:supp-texts} illustrates the edits
and perturbations applied to the reports. \cref{tab:supp-rarity} quantifies the frequency of the original and edited
label combinations in the CT-RATE training split. All edited and perturbed reports were fixed before any CT was generated.

\begin{table*}[t]
\centering\footnotesize\setlength{\tabcolsep}{4pt}
\caption{\textbf{Cohort, edits and perturbations.} Counts refer to cases and patients, not generated volumes.
The CTs of edited and perturbed reports are generated at the seeds of their original report. Swap generates no CT: it pairs
each CT with the labels (Report Agreement) or the CT (Eq.~(1)) of another case.}
\label{tab:supp-sets}
\begin{tabularx}{\linewidth}{@{}l l >{\raggedright\arraybackslash}X l@{}}
\toprule
Set & Cases (patients) & Construction & Used by \\
\midrule
Cohort & 940 (782) & one reconstruction per scan of the Text2CT evaluation list (1{,}817): soft-tissue kernel when
available (875), then the finest slice spacing; Findings and Impression; 18 CT-RATE labels (effusion 189, cardiomegaly 164) & all tests \\
Effusion edits & 567 (479) & 396 additions, 171 removals; 5 of the 572 reports with an effusion statement excluded, whose
statement cannot be separated from another finding & C1--C3 \\
Heart edits & 120 (120) & 50 enlargements, 70 normalisations; one case per patient among the effusion cases, drawn before
generation; normalisations exclude reports describing a dilated chamber & C1, C3 \\
Swap & 940 (782) & fixed derangement $j(i)$ of the 940 reports (seed 0); another patient in 939 cases & Report Agreement, Report Use (Eq.~(1)) \\
Scramble & 940 (782) & words of Findings and Impression permuted (seed 0), headings removed & Report Use \\
Off-topic & 940 (782) & passage of \emph{Moby-Dick} (Project Gutenberg \#2701) from a random start (seed 0), same word count & Report Use \\
\bottomrule
\end{tabularx}
\end{table*}

\begin{table*}[t]
\centering\footnotesize\setlength{\tabcolsep}{4pt}
\caption{\textbf{Examples of edits and perturbations.} Edits replace every statement about the edited finding and keep
the other findings, also within the same sentence. Replacement wording is drawn from CT-RATE
reports: the removal sentence occurs verbatim in 17 cohort reports, the addition follows the wording of 79 reports
describing bilateral effusion, and the enlargement and normalisation sentences occur in 71 and 124 reports. Swap: the donor report.}
\label{tab:supp-texts}
\begin{tabularx}{\linewidth}{@{}l >{\raggedright\arraybackslash}X >{\raggedright\arraybackslash}X@{}}
\toprule
Edit / perturbation & Original report (excerpt) & Modified report (excerpt) \\
\midrule
Effusion Add & No pleural or pericardial effusion was detected. & There is bilateral pleural effusion. No pericardial effusion was detected. \\
Effusion Remove & A free pleural effusion measuring 1\,cm in thickness is observed on the left. & No pleural effusion was detected. \\
Heart Enlarge & Mediastinal main vascular structures, heart contour, size are normal. & Mediastinal main vascular structures are normal. Heart size increased. \\
Heart Normalise & Coronary atherosclerosis and cardiomegaly. & Coronary atherosclerosis. Heart contour and size are normal. \\
\midrule
Swap & Findings: Trachea and both main bronchi are normal. No occlusive pathology was detected \ldots & Findings: Trachea and main bronchi are open. Right upper-bilateral lower paratracheal, \ldots \\
Scramble & (as for Swap) & be There both observed given. and the increased upper descending or both anterior \ldots \\
Off-topic & (as for Swap) & for a boat, and with his bandaged mouth seek to gag in death the vital jaw of the whale. \ldots \\
\midrule
Rewording, effusion stated & Bilateral pleural effusion is observed. & Pleural fluid is seen bilaterally. \\
Rewording, effusion negated & No pleural or pericardial effusion was detected. & Pleural and pericardial effusion were not detected. \\
Rewording, heart enlarged & Mild cardiomegaly. & Slightly increased heart size. \\
Rewording, heart normal & Heart contour, size is normal. & Cardiac dimensions and contour are within normal limits. \\
\bottomrule
\end{tabularx}
\end{table*}

\begin{table}[t]
\centering\footnotesize\setlength{\tabcolsep}{4pt}
\caption{\textbf{Rarity of the edited label combinations in the CT-RATE training split.} Labels: the 18 abnormality
labels that CT-RATE extracts from each report; an edited report keeps the labels of its original report, with the
edited finding set to the requested state. Training split: 24{,}128 scans of 20{,}000 patients, with 6{,}514 distinct
label combinations. Last row: the same counts for all 940 unedited cohort reports, as a reference; 30\% of real
reports already match no training scan, so a rare combination is not by itself unrealistic.}
\label{tab:supp-rarity}
\begin{tabular}{@{}lccc@{}}
\toprule
 & & Median training scans & No matching \\
 & & with the same labels, & training scan, \\
Edit & Cases & original $\to$ edited & original $\to$ edited \\
\midrule
Effusion Add & 396 & 10.5 $\to$ 1 & 17\% $\to$ 41\% \\
Effusion Remove & 171 & 0 $\to$ 0 & 64\% $\to$ 63\% \\
Heart Enlarge & 50 & 7 $\to$ 0 & 16\% $\to$ 62\% \\
Heart Normalise & 70 & 0 $\to$ 0 & 56\% $\to$ 53\% \\
\midrule
\begin{tabular}[t]{@{}l@{}}\textit{All 940 cohort}\\\textit{reports (reference)}\end{tabular} & \textit{940} & \textit{4} & \textit{30\%} \\
\bottomrule
\end{tabular}
\end{table}

\begin{figure*}[t]
\centering
\includegraphics[width=\linewidth,trim=0 274.32bp 0 0,clip]{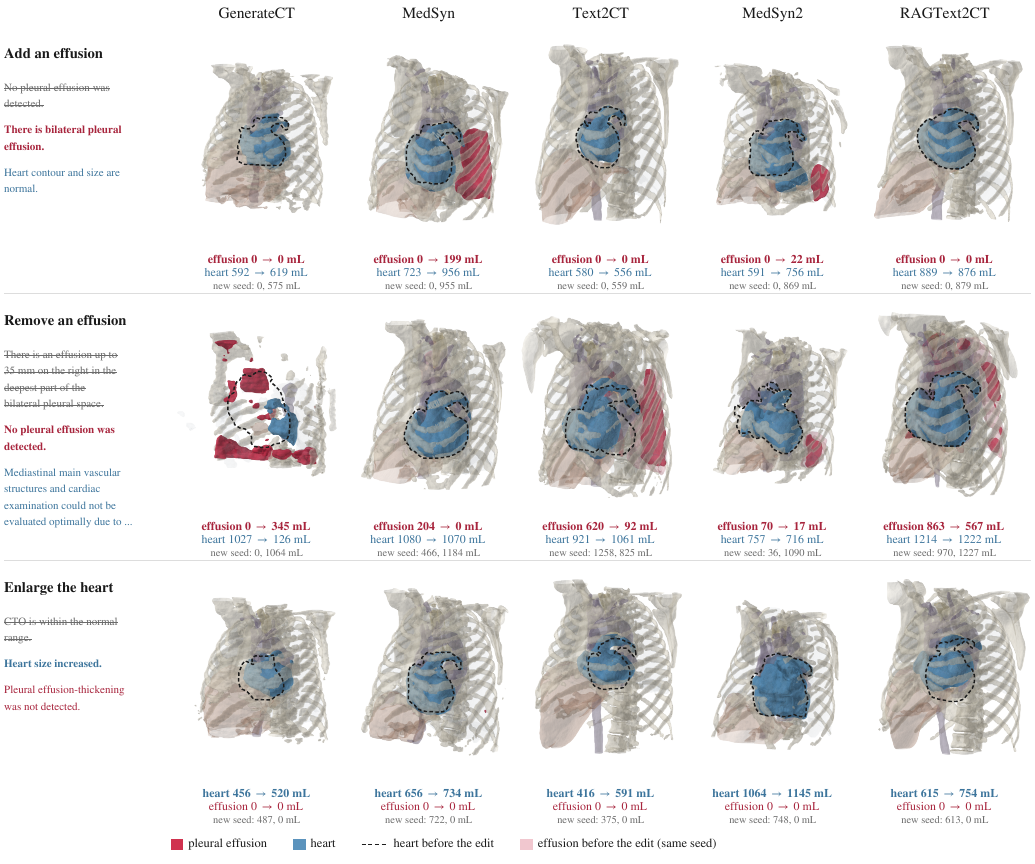}\\[-2pt]
\includegraphics[width=\linewidth,trim=0 0 0 268.56bp,clip]{fig_3d_cases/edit_atlas_3d.pdf}
\caption{\textbf{Two edits in 3D, for the five generators.} Rows: addition of an effusion (the case of Fig.~3) and
enlargement of the heart. Each cell shows the structures segmented by TotalSegmentator in the
CT of the edited report, at the seed of the case: heart (blue) and pleural effusion (crimson) solid; bones, vessels
and liver as context. The heart (dashed) and the effusion (pale) of the CT of the original report at the same seed are
overlaid. Left: the replaced sentence (struck through), its replacement (bold) and a sentence the edit leaves
unchanged. Below each cell: volume of the edited finding (bold) and of the other one, before $\to$ after; in grey, the
two volumes at another seed of the original report. Single cases do not estimate frequencies (Tab.~2).}
\label{fig:supp-3d}
\end{figure*}

\paragraph{Selection of qualitative examples.}\label{sec:supp-case} Let $q_{m,i}$ denote the percentile rank (midrank)
of the change in effusion volume for case $i$, $\Delta V_{m,i}$ (mL after the edit minus mL with the original report,
at the first seed), among the 396 additions evaluated with generator $m$. The eligible set $E$ contains 38 additions
involving a single sentence replacement, with complete CTs and segmentation masks for all five generators and the
corresponding real CT. The case shown in Fig.~3 is selected as
\begin{equation}
i^\star=\operatorname*{arg\,min}_{i\in E}\sum_{m=1}^{5}\bigl|q_{m,i}-0.5\bigr|.
\end{equation}
The rule
was fixed before visual inspection and selects case valid\_390\_a\_1 (CT-RATE volume names). \cref{fig:supp-3d} shows this case in 3D for the
five generators, together with a heart enlargement (valid\_324\_a\_2, selected by the same principle on the change in
heart volume). Fig.~1 shows another addition
(valid\_470\_a\_1), generated by one generator at two of its three seeds, chosen by hand to illustrate a strong (394\,mL)
and a weak (59\,mL) response to the same edit; the original report gives no effusion at any seed.

\subsection{Generators}
\label{sec:supp-generators}

\cref{tab:supp-contracts} lists, for each generator, the text encoder, training text, additional conditioning and
sampling settings. We consider as text-to-CT generators those whose only input is the report.
RAGText2CT qualifies: its pipeline takes only the report as input and retrieves from the report the anatomical mask
that conditions the generation. MedSyn2's optional masks would be a second input, so we do not use them.

\begin{table*}[t]
\centering\footnotesize\setlength{\tabcolsep}{3.5pt}
\caption{\textbf{Generators as run.} Every generator receives ``Findings: \ldots{} Impression: \ldots'' of the
report, whatever its training text; Additional conditioning: inputs besides the report. Reports longer than the window: among the 567 effusion reports. Text2CT: preprint checkpoint, the only one released. RAGText2CT retrieves from the 47{,}149 training
reports, as the 27{,}514-volume bank of its paper is not released.}
\label{tab:supp-contracts}
\begin{tabularx}{\linewidth}{@{}l >{\raggedright\arraybackslash}p{0.15\linewidth} c >{\raggedright\arraybackslash}p{0.13\linewidth} >{\raggedright\arraybackslash}p{0.12\linewidth} >{\raggedright\arraybackslash}X l@{}}
\toprule
Generator & Text encoder, window & \begin{tabular}[b]{@{}c@{}}Reports longer\\than the window\end{tabular} & Training text & Additional conditioning & Sampling (released pipeline) & Training data \\
\midrule
GenerateCT & T5, 256 tokens & 528 & age, sex, Impression & none & MaskGIT, 18 steps, guidance 5; diffusion super-resolution, 25 steps & CT-RATE \\
MedSyn & CXR-BERT, 2{,}048 positions, first 192 kept & 0 & UPMC reports & none & authors' diffusion sampler & UPMC \\
Text2CT & 3D CLIP, 512 tokens & 93 & Findings, Impression & none & rectified flow, 30 steps, guidance 5 & CT-RATE \\
MedSyn2 & CXR-BERT, 2{,}048 positions, first 192 kept & 0 & sections joined by a separator & none (masks not used) & midpoint solver & CT-RATE \\
RAGText2CT & 3D CLIP, 512 tokens & 93 & Findings, Impression & top-1 retrieved anatomical mask & as Text2CT; ControlNet scale 1 & CT-RATE \\
CTFlow (held out) & CXR-BERT with CT-CLIP weights, 128 tokens & -- & sex, age, report & none & autoregressive latent flow matching over slice blocks & CT-RATE \\
\bottomrule
\end{tabularx}
\end{table*}

\paragraph{Generation seeds.} Each case is assigned three deterministic seeds, shared across generators. The same seed gives different noise in
different architectures, so seeds pair CTs only within a generator. Within each
generator, the original and edited reports use the same seed for each paired generation. Credibility measures use only
the first seed. For case $i$ with identifier $\mathrm{id}_i$, the seed for repetition $k\in\{1,2,3\}$ is
\begin{equation}
s_{ik}=1+\bigl(h(\kappa_{ik})\bmod(2^{31}-1)\bigr),
\end{equation}
where $\kappa_{ik}$ is $\mathrm{id}_i$ for $k=1$, and $\mathrm{id}_i$ followed by \texttt{\#r1} or \texttt{\#r2} for
$k=2$ or $3$, respectively. The function $h$ converts the first 15 hexadecimal digits of the string's SHA-256 digest
to an integer. MedSyn2 restores the random state stored in its checkpoint; its generation seed is therefore set after
checkpoint loading. Identical text and seed gave bit-identical volumes in every case we checked.

\subsection{Measures}
\label{sec:supp-measures}

\paragraph{Confidence intervals.} Unless stated otherwise, we compute 95\% confidence intervals using 2{,}000 bootstrap
resamples of patients (1{,}000 for Report Agreement). Patients are sampled with replacement, retaining all cases and seeds of each sampled patient.
We recompute the metric for each resample and use the 2.5th and 97.5th percentiles as interval bounds.
Diversity reports the standard deviation over its 20 splits of the patients, and the FD is computed once.

\paragraph{Realism: Fr\'echet distance.} The FD of Realism is computed from RadImageNet ResNet50 features of 24 evenly spaced slices per
plane, after resampling CTs to a 0.7\,mm isotropic grid and applying the lung window. We compute FD separately for axial, coronal, and sagittal slices and average the three
values. FD depends on sample size, and its real reference needs real CTs disjoint from those it is compared with. We
therefore split the 782 patients into A (200 patients, one CT each) and B (the other 582 patients, 692 CTs). We
compute every FD against B, from 200 CTs: the real CTs of A for the real reference, the same CTs blurred for the null
control, and the CTs generated from the reports of A for each generator. \cref{fig:supp-known}a gives its response to known image
changes.

\paragraph{Realism and Diversity: precision, recall and coverage.} Precision measures Realism; recall and coverage
measure Diversity (Sec.~3.1). Unlike the FD, all three use Euclidean distances between
unit-normalised 3D \mbox{Pillar-0} features, with
neighbourhood radii defined by the fifth nearest neighbour. We randomly split patients into two halves and compare
synthetic CTs generated from reports of one half with real CTs from the other. We average each metric over 20 random
splits. The real reference uses the same splits with real CTs in both halves.

\paragraph{Report Use.}\label{sec:supp-reportuse} We check the scale of Eq.~(1) by computing it on real CTs. The
numerator becomes the mean distance between each of 200 real CTs and the same CT after one of the known changes of
\cref{fig:supp-known}, such as noise or blur. The denominator becomes the mean distance
between each of the 940 real CTs and the real CT of another case. \cref{fig:supp-known}b gives the resulting $r$
for each change: it is 0 for the identity, grows with the strength of noise and blur (0.20, 0.55 and 0.93 for a blur
of 1, 2 and 4\,mm), and reaches 1, the distance from the CT of another case, only when the slices are shuffled
(1.03), \ie a change that destroys the 3D anatomy.

\paragraph{Report share.} The report share is the fraction of feature variance, over cases and seeds, explained
by the report (Sec.~5, \cref{tab:diagnose}). For case $i$, let $\phi_{ik}$ be the unit-normalised Pillar-0 features
of the CT generated from its unedited report at seed $k$, $\bar\phi_i$ their mean over the $K=3$ seeds, and
$\bar\phi$ the mean over the $n$ cases. The report share is the one-way intraclass correlation (ICC)
\begin{equation}
\begin{aligned}
\mathrm{ICC}&=\frac{\mathrm{MS_b}-\mathrm{MS_w}}{\mathrm{MS_b}+(K-1)\,\mathrm{MS_w}},\\
\mathrm{MS_b}&=\frac{K}{n-1}\sum_i\|\bar\phi_i-\bar\phi\|^2,\\
\mathrm{MS_w}&=\frac{1}{n(K-1)}\sum_{i,k}\|\phi_{ik}-\bar\phi_i\|^2,
\end{aligned}
\end{equation}
where $\|\cdot\|$ is the Euclidean norm, $\mathrm{MS_b}$ the mean square between cases (how much the CTs of different
reports differ) and $\mathrm{MS_w}$ the mean square within cases (how much the CTs of the same report differ across
seeds). The report share is 1 when the seeds of a report give the same features, and 0 when CTs of different reports
differ no more than CTs of different seeds. We use the 100 effusion-edit cases of the Lite suite, on which CTFlow was also run. Values
range from 13\% (MedSyn2) to 64\% (GenerateCT), and are 13\% for CTFlow (\cref{tab:external}).

\paragraph{Utility.} We use five patient-disjoint folds, stratified by pleural effusion, and $L_2$-regularised
logistic regression
($C=1$) on standardised Pillar-0 features. For each finding, the classifier is trained on synthetic CTs from four folds
and evaluated on real CTs from the remaining fold. We pool the out-of-fold predictions, compute AUROC for each finding,
and average across the 18 findings. The reference trains the same classifier on the real CTs of the same folds.

\paragraph{Effusion Add and Remove.} \cref{tab:supp-c1} splits the net effect into transitions between the CTs of
the original and the edited report at the same seed. Let $p_{ab}$ be the share of case--seed pairs in state $a$
before the edit and $b$ after it, with 1 the requested state. The net effect is $p_{01}-p_{10}$: pairs that do not
change count zero, whatever the baseline prevalence.

\begin{table}[t]
\centering\footnotesize\setlength{\tabcolsep}{2.2pt}
\caption{\textbf{Transitions behind C1}, \% of case--seed pairs; 1 = state requested by the edit (TotalSegmentator, $\geq5$\,mL over both lungs); $p_{ab}$: state $a$ with the original report, $b$ with the edited one. Net $=p_{01}-p_{10}$, the value of Tab.~2.}
\label{tab:supp-c1}
\input{assets/proposals/supplement/tables/c1_transitions.tex}
\end{table}

\begin{figure}[t]
\centering
\includegraphics[width=\linewidth]{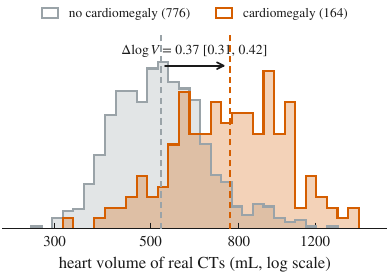}
\caption{\textbf{Log heart volume of the 940 real CTs} with and without report-labelled cardiomegaly; dashed: means.
Their difference is the C1 heart reference.}
\label{fig:supp-heart-ref}
\end{figure}

\paragraph{C1 on the heart: reference.} Let $V$ be the heart volume (TotalSegmentator) of each of the 940 real CTs,
all above 100\,mL, and $\ell$ its cardiomegaly label. The reference of Heart Enlarge and Heart Normalise is the
difference in mean log volume between the 164 real CTs with cardiomegaly and the 776 without (\cref{fig:supp-heart-ref}):
\begin{equation}
\Delta\log V_{\text{real}}=\overline{\log V}_{\ell=1}-\overline{\log V}_{\ell=0}=0.37\ [0.31,0.42].
\end{equation}
A heart case is included in C1 when the heart is segmented ($\geq100$\,mL) in all six of its CTs (original and edited
report, three seeds). This holds for 48 of the 50 enlargements and 58 of the 70 normalisations with Text2CT, 41 and
52 with GenerateCT, 50 and 65 with RAGText2CT, and all with MedSyn and MedSyn2.

\paragraph{C2: Negated Findings.} Containment is measured on the effusion edits only; the 120 heart edits serve to
test whether Target Response and Reliability depend on the edited finding (Sec.~5). For each effusion addition and
each other finding that the report negates (every sentence mentioning it contains a negation), NV-Reason-CT checks
whether the finding is present (P(yes) $\geq0.5$) in the CTs of the original and of the edited report, at the same
seed. Negated Findings (NF) is the mean rise in these detections after the edit, over seeds and findings, minus the
rise expected from the classifier's excess (\cref{sec:supp-excess}):
\begin{equation}
\mathrm{NF}=\frac{1}{|P|}\sum_{(i,g)\in P}\Bigl[\frac1K\sum_{k}\bigl(c_g(x_{ik}(T))-c_g(x_{ik})\bigr)-\widehat\Delta_{\mathrm{eff}}\,e_g\Bigr],
\end{equation}
where $P$ is the set of pairs of an addition $i$ and a finding $g$ negated in $t_i$, $c_g(x)=1$ when NV-Reason-CT detects $g$
in $x$, $\widehat\Delta_{\mathrm{eff}}$ is the classifier's net effect on the effusion and $e_g$ its excess for
$g$. Edits beyond the text window are left out. Without the correction, NF is 0.06, 0.15, 0.03, 0.08 and 0.01 for
GenerateCT, MedSyn, Text2CT, MedSyn2 and RAGText2CT (0.02, 0.09, 0.03, 0.04 and 0.01 with it, Tab.~2): MedSyn remains
the worst and RAGText2CT the best, and only GenerateCT and Text2CT swap places. NF $=0$ means no net rise: it does not
exclude negated findings already detected before the edit, or rises and falls that cancel.

\begin{figure}[t]
\centering
\includegraphics[width=\linewidth]{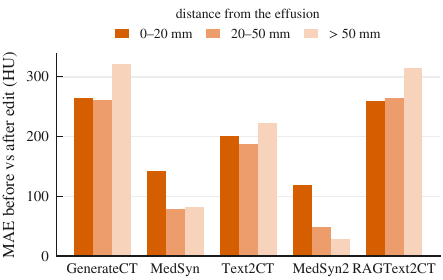}
\caption{\textbf{Same-seed change by distance from the effusion.} Mean absolute difference (HU) between the CTs of the
original and edited report; first seed, edits beyond the text window excluded.}
\label{fig:supp-bands}
\end{figure}

\paragraph{C2: the region beyond 50\,mm.} For case $i$, $\mathcal R_i$ contains the body voxels more than 50\,mm
away from any effusion segmented in its six CTs (original and edited report, three seeds), and $g$ is the vector of
their intensities. An effusion legitimately displaces the lung next to it, by about its own thickness (\cref{fig:supp-bands}), and 50\,mm covers this displacement for most real effusions: of the 2{,}858 CT-RATE training scans labelled with an effusion,
1{,}348 report its thickness, with a median of 30\,mm, and 80\% of these are at most 50\,mm thick. On the segmentation masks of the 940 cohort CTs, the 238 effusions of at least 5\,mL have a median thickness of 25\,mm and 83\% are at most 50\,mm thick (Spearman 0.82 with the thickness stated in the report, for the 105 that state it).

\paragraph{C2: estimating $\theta$.}\label{sec:supp-theta} Eq.~(3) compares the mean shift induced by the edit, $\|\delta_i\|^2$ where $\delta_i=\mu_{g,i}^{(1)}-\mu_{g,i}^{(0)}$, with the variability due to the seed,
$v_i^{(0)}+v_i^{(1)}$ (we drop the index $g$). We estimate both from $K=3$ seeds per case. Sample variances are
unbiased estimates of the variability. The squared observed mean shift, instead, overestimates $\|\delta_i\|^2$,
because it includes the uncertainty due to the finite number of seeds.

To correct this overestimate, let $d_{ik}$ be the difference between the edited and the original CT generated with
the same seed $k$, $\bar d_i$ their mean and $S_{d,i}$ their sample covariance. If $c_i$ is the trace of the
covariance between the original and the edited CT, then
\begin{equation}
\begin{aligned}
\mathbb E\|\bar d_i\|^2&=\|\delta_i\|^2+\frac{v_i^{(0)}+v_i^{(1)}-2c_i}{K},\\
\mathbb E\operatorname{tr}S_{d,i}&=v_i^{(0)}+v_i^{(1)}-2c_i.
\end{aligned}
\end{equation}
The term $\operatorname{tr}S_{d,i}/K$ therefore estimates the contribution of the seed variability to the squared
observed shift. Subtracting it gives
\[
\widehat A_i=\|\bar d_i\|^2-\frac{\operatorname{tr}S_{d,i}}{K},
\]
an unbiased estimate of $\|\delta_i\|^2$, whatever the correlation introduced by the shared seed. We then estimate
$\theta$ as
\begin{equation}
\hat\theta_g=\frac{\sum_i\widehat A_i}{\sum_i\bigl(\hat v_i^{(0)}+\hat v_i^{(1)}\bigr)}.
\end{equation}
All squared norms and traces are normalised by the number of voxels of $\mathcal R_i$. Confidence intervals are
computed by bootstrap over patients.

To read the scale: for a single scalar readout with equal variances under the two reports, $\theta=d^2/2$, with $d$
Cohen's standardised difference~\cite{Cohen_1988_Power}, so $\theta=1$ corresponds to $|d|\approx1.4$, a large shift relative to the seed variability. $\theta$ measures a
shift of the mean only: a small $\theta$ does not exclude changes in the variance or in the distribution of the
intensities. The simulation with a known shift (\cref{tab:supp-theta-sanity}) shows that the
estimator recovers $\theta$ at every level of correlation $\rho$ considered. The paired MAE, instead, falls tenfold
as $\rho$ grows even with no shift, which shows its dependence on the correlation introduced by the shared seed.

\begin{table}[t]
\centering\footnotesize\setlength{\tabcolsep}{4pt}
\caption{\textbf{$\hat\theta$ does not depend on the seed coupling; the paired MAE does.} Mean of 500 simulations of
567 cases with three seeds: $x_{0k}=\mu_i+\sigma\epsilon_k$,
$x_{1k}=\mu_i+\delta+\sigma(\rho\epsilon_k+\sqrt{1-\rho^2}\eta_k)$, $\sigma=1$, true $\theta=\delta^2/2\sigma^2$.}
\label{tab:supp-theta-sanity}
\begin{tabular}{@{}lc cccc@{}}
\toprule
 & & \multicolumn{4}{c}{seed coupling $\rho$} \\ \cmidrule(l){3-6}
$\delta$ (true $\theta$) & & 0 & 0.5 & 0.9 & 0.99 \\
\midrule
0 (0) & $\hat\theta$ & 0.001 & 0.000 & 0.000 & 0.000 \\
 & MAE & 1.127 & 0.797 & 0.356 & 0.113 \\
1 (0.5) & $\hat\theta$ & 0.499 & 0.502 & 0.500 & 0.501 \\
 & MAE & 1.399 & 1.167 & 1.004 & 1.000 \\
$\sqrt2$ (1) & $\hat\theta$ & 1.002 & 1.004 & 1.003 & 1.001 \\
 & MAE & 1.650 & 1.487 & 1.414 & 1.414 \\
\bottomrule
\end{tabular}
\end{table}

\paragraph{C3: Seed Agreement.}\label{sec:supp-c3def} We assess whether edit success is consistent across the
$K=3$ seeds. For each case and seed, we assign a binary success outcome:
\begin{itemize}\setlength{\itemsep}{0pt}
\item Effusion: the edited CT is in the requested state (effusion present after an addition, absent after a removal). Only eligible cases are included: those whose CTs from the original report are in the opposite state at all three seeds.
\item Heart: the heart volume increases after an enlargement or decreases after a normalisation, on the heart cases included in C1.
\end{itemize}
Seed Agreement is measured by Fleiss' $\kappa$~\cite{Fleiss_1971_Kappa}, treating seeds as raters and success or
failure as the two outcome categories. For $n$ cases, let $n_i$ denote the number of seeds at which the edit succeeds
for case $i$. The observed pairwise agreement is
\[
\bar P=\frac{1}{n}\sum_i\frac{n_i(n_i-1)+(K-n_i)(K-n_i-1)}{K(K-1)}.
\]
The overall success proportion $\bar p$ determines the agreement expected by chance, $P_e$:
\[
\bar p=\frac{\sum_i n_i}{nK},\qquad P_e=\bar p^2+(1-\bar p)^2,\qquad \kappa=\frac{\bar P-P_e}{1-P_e}.
\]
Provided both outcome categories occur, $\kappa=1$ when every case has the same outcome across all three seeds,
whether success or failure. Thus, Seed Agreement measures consistency, not the success rate (read with C1). When
success or failure dominates, $\kappa$ can be unstable; it is undefined when all outcomes are identical.
\cref{fig:supp-c3} therefore also reports the proportions of cases with success at all three seeds, at one or two
seeds, or at none.

%% file: assets/proposals/supplement/tables/published.tex
\begin{tabularx}{\linewidth}{@{}l >{\raggedright\arraybackslash}X >{\raggedright\arraybackslash}X >{\raggedright\arraybackslash}p{0.30\linewidth}@{}}
\toprule
Work & Compared baselines & Evaluation metrics & Partly met in Tab.~1 \\
\midrule
GenerateCT~\cite{Hamamci_2024_GenerateCT} & Imagen, Stable Diffusion, Phenaki & FID, FVD (I3D, CT-Net), CLIP score & Real-CT reference: reported for training pairs only \\
MedSyn~\cite{Xu_2024_MedSyn} & WGAN, $\alpha$-GAN, HA-GAN, Medical Diffusion & FID, MMD, airway volume & Multiple seeds: shown with anatomy masks, not quantitatively evaluated \\
Text2CT~\cite{Molino_2025_Text2CT} & GenerateCT, MedSyn, Report2CT & FID (2.5D, 3D), AUC, precision, anatomical realism, report consistency & Multiple seeds: FID averaged over 10 seeds \\
RAGText2CT~\cite{Molino_2026_RAGText2CT} & GenerateCT, MedSyn, Text2CT, Report2CT, MAISI & FID (2.5D, 3D), AUC, precision, Dice, HD95 & Multiple seeds: 5 seeds used for significance tests \\
MedSyn2~\cite{MedSyn2_2026} & GenerateCT, MedSyn, Text2CT & FID, CLIP scores (T2I, I2I), accuracy, precision, AUROC, F1 & Same-seed edits: qualitative only; multiple seeds: mask-conditioned examples; utility: mask-only segmentation \\
CTFlow~\cite{Wang_2025_CTFlow} & Imagen, Stable Diffusion, Phenaki, GenerateCT & FID, FVD, CLIP score & -- \\
\bottomrule
\end{tabularx}

%% file: assets/proposals/supplement/tables/c1_transitions.tex
\begin{tabular}{@{}l rrrrc rrrrc@{}}
\toprule
 & \multicolumn{5}{c}{Effusion Add} & \multicolumn{5}{c}{Effusion Remove} \\ \cmidrule(lr){2-6}\cmidrule(l){7-11}
Generator & $p_{00}$ & $p_{01}$ & $p_{10}$ & $p_{11}$ & Net & $p_{00}$ & $p_{01}$ & $p_{10}$ & $p_{11}$ & Net \\
\midrule
GenerateCT & 59.8 & 26.1 & 5.9 & 8.2 & \textbf{20.2} & 13.1 & 18.1 & 9.2 & 59.6 & \textbf{9.0} \\
MedSyn & 38.1 & 56.6 & 0.0 & 5.3 & \textbf{56.6} & 33.3 & 40.7 & 0.2 & 25.7 & \textbf{40.5} \\
Text2CT & 88.6 & 4.5 & 2.0 & 4.9 & \textbf{2.4} & 59.3 & 23.6 & 2.3 & 14.8 & \textbf{21.2} \\
MedSyn2 & 37.6 & 47.1 & 0.3 & 15.1 & \textbf{46.8} & 42.1 & 27.5 & 0.0 & 30.4 & \textbf{27.5} \\
RAGText2CT & 87.2 & 6.0 & 1.2 & 5.6 & \textbf{4.8} & 61.2 & 21.4 & 3.5 & 13.8 & \textbf{17.9} \\
\midrule
Pairs & \multicolumn{5}{c}{396 cases $\times$ 3 seeds} & \multicolumn{5}{c}{171 cases $\times$ 3 seeds} \\
\bottomrule
\end{tabular}

%% file: appendix/e02_evaluators.tex
\section{Evaluators}
\label{sec:supp-validation}

\begin{figure*}[t]
\centering
\includegraphics[width=\linewidth]{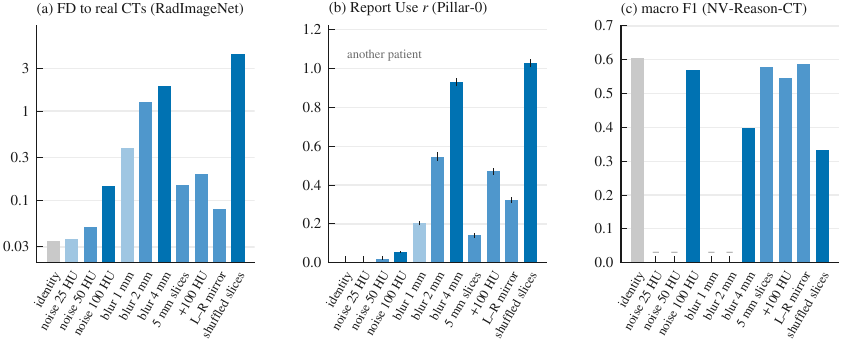}
\caption{\textbf{Evaluators under known changes} of 200 real CTs of 200 patients (NV-Reason-CT: 100), darker for
stronger changes; noise and blur are Gaussian, with standard deviation in HU and mm; --: not run. (a) FD to the real
CTs of other patients (log scale): the identity gives the real reference of Realism (0.035), a 4\,mm blur its null
control (1.91). (b) Report Use, Eq.~(1), between each CT and its changed version: 0 for the identity, and as far as
another patient (dashed) only when the slices are shuffled; bars: 95\% intervals (bootstrap over patients). (c) Macro F1 of
NV-Reason-CT over the 18 findings.}
\label{fig:supp-known}
\end{figure*}

\cref{tab:supp-evaluators} lists the evaluators and what each is used for. On the 940 real CTs, each matches its
published value or the findings of the reports, and responds as expected to known image changes
(\cref{fig:supp-known}).

\begin{table*}[t]
\centering\footnotesize\setlength{\tabcolsep}{4pt}
\caption{\textbf{Evaluators}, used as released, and their check on the 940 real CTs of the cohort (in parentheses:
published value; Pillar-0: value of the MedSyn2 paper). Responses to known image changes: \cref{fig:supp-known}.}
\label{tab:supp-evaluators}
\begin{tabularx}{\linewidth}{@{}l >{\raggedright\arraybackslash}X >{\raggedright\arraybackslash}X >{\raggedright\arraybackslash}X@{}}
\toprule
Evaluator & Output & Used for & On the 940 real CTs \\
\midrule
RadImageNet~\cite{Mei_2022_RadImageNet} & 2.5D slice features & Realism (FD) & FD between patient-disjoint subsets: 0.035 \\
Pillar-0~\cite{Agrawal_2025_Pillar0} & 3D CT and report embeddings & precision, recall, coverage; Report Use; Utility & report--CT cosine $\times100$: 27.1 (26.4) \\
NV-Reason-CT~\cite{Myronenko_2026_NVReasonCT} & P(yes) of the 18 findings & Report Agreement; Negated Findings & macro AUROC: 0.85 (0.87) \\
TotalSegmentator~\cite{Wasserthal_2023_TotalSegmentator} & effusion and heart masks & C1; C3; Distant Anatomy; Pre-Flight Checks & effusion: Youden's $J$ 0.86; heart volume for cardiomegaly: AUROC 0.83 \\
\bottomrule
\end{tabularx}
\end{table*}

CT-CLIP~\cite{Hamamci_2024_CTRATE}, used in earlier papers, reaches a macro AUROC of 0.68 (published 0.67) on the
same CTs, against 0.85 for NV-Reason-CT, which we therefore use. Under the left--right mirror, both finding
evaluators read the effusion on the other side, in 55 of 55 CTs (NV-Reason-CT) and 50 of 50 (TotalSegmentator), among
the 71 real CTs with a one-sided effusion: they read laterality from the image.

\paragraph{Which findings are edited.} Both finding evaluators must detect the finding on the 940 real CTs against
the CT-RATE labels (AUROC $\geq0.80$ or Youden's $J\geq0.50$; Sec.~4.3). Pleural effusion (NV-Reason-CT AUROC 0.97;
TotalSegmentator $J$ 0.86) and cardiomegaly (0.92; heart volume AUROC 0.83) pass. Lung nodules do not (0.799; 0.53).

\paragraph{The classifier's excess.}\label{sec:supp-excess} On the 940 real CTs, we measure the association between
the report's effusion label $\ell_{\mathrm{eff}}$ and each other finding $g$ twice: once with $g$ as detected by
NV-Reason-CT ($c_g$), once with its report label ($\ell_g$). The classifier's excess is the difference between these
two estimates:
\begin{equation}
\begin{aligned}
e_g={}&\bigl[\Pr(c_g{=}1\mid\ell_{\mathrm{eff}}{=}1)-\Pr(c_g{=}1\mid\ell_{\mathrm{eff}}{=}0)\bigr]\\
&-\bigl[\Pr(\ell_g{=}1\mid\ell_{\mathrm{eff}}{=}1)-\Pr(\ell_g{=}1\mid\ell_{\mathrm{eff}}{=}0)\bigr].
\end{aligned}
\end{equation}
For example, for cardiomegaly $e_g=0.14$ [0.06, 0.22]: the classifier associates it with effusion more than the
reports do. Over the 17 findings, $e_g$ ranges from $-0.09$ to $0.15$, and its interval excludes zero for
seven of them. The excess is an association measured on real CTs, not a causal effect of the effusion on the
classifier. Assuming that it transfers to edited CTs, an increase $\widehat\Delta_{\mathrm{eff}}$ in the detection rate
of effusion raises that of $g$ by about $\widehat\Delta_{\mathrm{eff}}\,e_g$, even when $g$ is unchanged. Negated
Findings subtracts this contribution, so it should be read under this assumption.

%% file: appendix/e03_results.tex
\section{Results}
\label{sec:supp-results}

\subsection{Tab.~2 with Intervals}
\label{sec:supp-main}

\cref{tab:supp-main-ci} gives the measures of Tab.~2 with their 95\% intervals.

\begin{table*}[tp]
\centering\footnotesize\setlength{\tabcolsep}{3pt}
\caption{\textbf{Tab.~2 with 95\% intervals} (bootstrap over patients, \cref{sec:supp-measures}). Grey: null control and
real or ideal value; --: no null control. Heart edits: 120 cases; other Control measures: 567 effusion edits. Realism:
one estimate on 200 synthetic \vs 692 real CTs, without interval; on all 940 CTs, with a bootstrap over patients, the
five generators keep this order and every pairwise difference excludes zero. $^\dagger$Mean $\pm$ one standard
deviation over 20 splits of the patients.}
\label{tab:supp-main-ci}
\input{assets/proposals/supplement/tables/main_ci.tex}
\end{table*}

\subsection{The Ideal Value of Report Use}
\label{sec:supp-reading}

Report Use compares the CTs generated at the same seed from a report and from its perturbation (Eq.~(1)): it measures how much the CT moves with the text, not whether the CT follows its meaning, and it fails at both ends. At $r=0$ the CT does not move: the perturbation is ignored. At $r\geq1$ the perturbation moves the CT as far as changing the case does, although the seed is shared; on real CTs, among the known changes only slice shuffling, which destroys
the 3D anatomy, reaches this distance (1.03, \cref{sec:supp-reportuse}). No reference fixes a correct value in
between: a real CT does not change with its report, and the task does not say how far a CT should move. We therefore take as ideal, by convention, the value farthest from both failures, $r=0.5$, and score Report Use as
\begin{equation}
s=1-2\,\bigl|\min(r,1)-0.5\bigr|,
\end{equation}
which is 0 at both failures and 1 at $r=0.5$. The Report Card places $s$ between its centre (0) and rim (1).

\subsection{C1: Within the Text Window and at \texorpdfstring{10\,mL}{10 mL}}
\label{sec:supp-c1}

\cref{tab:supp-effusion} repeats the effusion measures of Tab.~2 in two variants. Left: only the edits that fall within
the text window of each generator (Sec.~4.2): 419 of the 567 for GenerateCT, 566 for Text2CT and RAGText2CT, all for
MedSyn and MedSyn2. Right: all edits, with an effusion counted as present from 10\,mL instead of 5\,mL, to test the
choice of threshold. On the 940 real CTs, both thresholds agree equally well with the report labels (Youden's $J$
0.86). At both thresholds, the five generators rank in the same order on Effusion Add and on Effusion Remove.

\begin{table*}[tp]
\centering\footnotesize\setlength{\tabcolsep}{3pt}
\caption{\textbf{C1 and Seed Agreement on the effusion edits, within the text window and at 10\,mL.} Effusion Add and
Remove: net effect over three seeds (TotalSegmentator; effusion present from 5\,mL unless stated). Seed Agreement:
Fleiss' $\kappa_+$ of an addition's success; for GenerateCT, on the 230 of its 315 eligible additions within the window. 95\% intervals (bootstrap over patients).}
\label{tab:supp-effusion}
\input{assets/proposals/supplement/tables/effusion_window.tex}
\end{table*}

\subsection{C2: Distant Anatomy of RAGText2CT}
\label{sec:supp-distant}

RAGText2CT conditions each CT on the anatomical mask of the training CT that its report retrieves
(\cref{tab:supp-contracts}). An edit changes the report and can therefore change the retrieved CT.
\cref{tab:supp-rag-theta} splits its $\theta_+$ (1.79 in Tab.~2) by whether the edited report retrieves another
training CT (249 of the 396 additions) or the same one (147).

\begin{table*}[tp]
\centering\footnotesize
\begin{minipage}[t]{0.40\linewidth}\centering
\caption{\textbf{Distant Anatomy of RAGText2CT by retrieved training CT.} $\theta_+$ beyond 50\,mm (\cref{sec:supp-theta}), 95\% intervals (bootstrap over patients).}
\label{tab:supp-rag-theta}
\input{assets/proposals/supplement/tables/rag_theta.tex}
\end{minipage}\hfill
\begin{minipage}[t]{0.56\linewidth}\centering\setlength{\tabcolsep}{3pt}
\caption{\textbf{Rewording verdicts} (\cref{sec:supp-reword}). Readouts: effusion presence ($\geq5$\,mL) and
$\log(1+\mathrm{mL})$, heart $\log V$ (TotalSegmentator); effusion and cardiomegaly calls (NV-Reason-CT). --: cannot be
assessed (the edit that changes the meaning has no detectable effect). Bold: the verdict holds after Holm adjustment
over the 25 pairs.}
\label{tab:supp-reword}
\input{assets/proposals/supplement/tables/rewording.tex}
\end{minipage}
\end{table*}

\subsection{C3: Seed Agreement on All Four Edits}
\label{sec:supp-c3}

\begin{figure*}[tp]
\centering
\includegraphics[width=\linewidth]{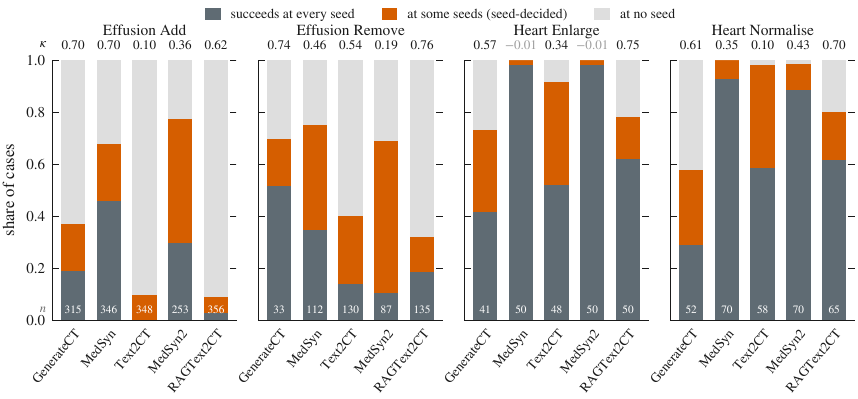}
\caption{\textbf{Seed Agreement on the four edits.} Bars: cases eligible at all three seeds ($n$ at the base), by
whether the edit succeeds at every seed, at some (seed-decided) or at none. Above: Fleiss' $\kappa$, in grey when at
least 95\% of the cases succeed at every seed or fail at every seed. In these cases $\kappa$ is unstable and can be close to 0
even when the seeds agree.}
\label{fig:supp-c3}
\end{figure*}

Tab.~2 gives Seed Agreement for Effusion Add. \cref{fig:supp-c3} extends it to Effusion Remove, Heart Enlarge and
Heart Normalise. Since $\kappa$ measures consistency rather than success, and is unstable when one outcome dominates
(\cref{sec:supp-c3def}), the figure also splits the eligible cases by whether the edit succeeds at every seed, at
some, where the seed decides the outcome, or at none.

\subsection{Findings AUROC on Edited CTs}
\label{sec:supp-edit-auroc}

\cref{tab:supp-edit-auroc} reports the AUROC of NV-Reason-CT against the labels of the edited reports, on the CTs
generated from the original and from the edited reports: over the 18 findings (macro AUROC) and for the edited finding
alone (effusion AUROC).

\begin{table*}[tp]
\centering\footnotesize\setlength{\tabcolsep}{4pt}
\caption{\textbf{Findings AUROC against the edited reports} (Fig.~6b), first seed, 567 effusion edits. Edited and
Original: CTs generated from the edited and from the original reports. Difference: Edited minus Original, with its 95\% interval (bootstrap over patients).}
\label{tab:supp-edit-auroc}
\input{assets/proposals/supplement/tables/edit_auroc_seed1.tex}
\end{table*}

\subsection{The Sentence and the Rest of the Report}
\label{sec:supp-sentence}

\begin{figure*}[tp]
\centering
\includegraphics[width=\linewidth]{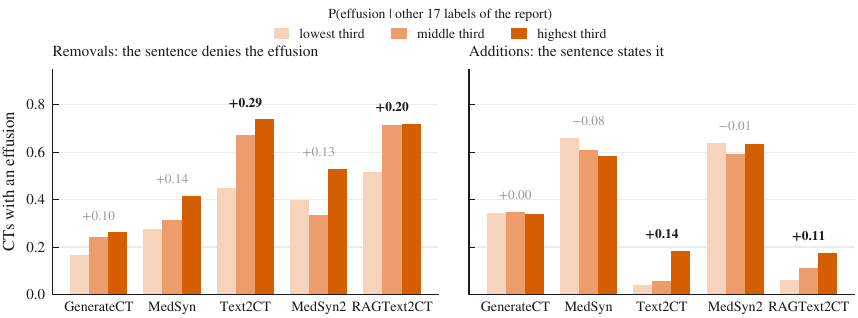}
\caption{\textbf{Share of CTs with an effusion after the edit, by third of $\hat P_i$} (TotalSegmentator, three seeds). $\hat P_i$: probability of an effusion predicted from the other 17 labels of the report.
Left: the 171 removals; right: the 396 additions. Numbers: highest minus lowest third, bold when its 95\% interval (bootstrap over patients) excludes zero. Text2CT after a removal: 45\% (lowest third) and 74\% (highest third).}
\label{fig:supp-terciles}
\end{figure*}

To remove or add a pleural effusion we always insert the same sentence, ``No pleural effusion was detected.'' or
``There is bilateral pleural effusion.''; only the rest of the report varies. We score each report $t_i$ by $\hat P_i$,
the probability of an effusion predicted from its own other 17 labels by a logistic regression without interaction
terms, fitted once on the CT-RATE training labels used for \cref{tab:supp-rarity}. Separately for removals and
additions, we cut the cases into thirds of $\hat P_i$; \cref{fig:supp-terciles} gives, for each generator and third,
the share of CTs (three seeds per case) with an effusion after the edit. If only the sentence mattered, the three bars
of a generator would have the same height. The difference between the highest and the lowest third excludes zero in
both directions for Text2CT and RAGText2CT, the two generators with the lowest Effusion Add in Tab.~2, and in neither
for GenerateCT, MedSyn and MedSyn2. Where the sentence has little effect, the rest of the report decides: editing one
sentence is not always enough to control one finding (Sec.~6).

\subsection{Rewording: an Invariance Test}
\label{sec:supp-reword}

Adding a finding tests whether the generator changes the CT in the requested direction; rewording a statement without
changing its meaning tests whether the result stays the same~\cite{Ribeiro_2020_CheckList}. We reworded the effusion
or heart statement of 79 reports, one per patient: 19 stating an effusion, 20 negating it, 20 describing an enlarged
heart and 20 a normal one. About half of the rewordings keep the term and change the syntax; the others replace it
with a clinical synonym (\cref{tab:supp-texts}). CTs are generated at the seed of the original report. For each
readout $h$, we compare two changes of the same statement: rewording it, with mean change $D_W$, and editing it to the
opposite meaning, with mean change $D_E$. A stated effusion is removed, a negated one is added, an enlarged heart is
normalised and a normal heart is enlarged. With a margin $\varepsilon$
fixed in advance (0.10 for presence, 0.05 for $\log V$, 0.25 for $\log(1+\mathrm{mL})$), a generator:
\begin{itemize}\setlength{\itemsep}{0pt}
\item \emph{follows the meaning} when rewording changes $h$ little, in absolute terms and relative to the edit: the
90\% interval of $D_W$ lies within $\pm\varepsilon$ (two one-sided tests~\cite{Schuirmann_1987_TOST}), the 95\%
interval of $D_E-D_W$ lies above zero, and the 95\% interval of $D_W/D_E$ lies below $1/2$;
\item \emph{depends on the wording} when the 95\% interval of $D_W$ excludes zero;
\item is \emph{inconclusive} otherwise, and \emph{cannot be assessed} when the edit has no effect.
\end{itemize}
\cref{tab:supp-reword} gives the verdict for each generator and readout; in bold, those that hold after Holm
adjustment~\cite{Holm_1979_Multiple} over the 25 pairs.

%% file: assets/proposals/supplement/tables/main_ci.tex
\begin{tabular}{@{}lcclllll@{}}
\toprule
Measure & Null & Real or ideal & GenerateCT & MedSyn & Text2CT & MedSyn2 & RAGText2CT \\
\midrule
\multicolumn{8}{@{}l}{\textit{C1 Target Response}} \\
Effusion Add$\uparrow$ & \textcolor{black!50}{0.00} & \textcolor{black!50}{1.00} & +0.20 {\scriptsize [0.16, 0.25]} & +0.57 {\scriptsize [0.52, 0.61]} & +0.02 {\scriptsize [0.01, 0.04]} & +0.47 {\scriptsize [0.43, 0.50]} & +0.05 {\scriptsize [0.03, 0.07]} \\
Effusion Remove$\uparrow$ & \textcolor{black!50}{0.00} & \textcolor{black!50}{1.00} & +0.09 {\scriptsize [0.03, 0.16]} & +0.41 {\scriptsize [0.34, 0.47]} & +0.21 {\scriptsize [0.16, 0.27]} & +0.27 {\scriptsize [0.23, 0.32]} & +0.18 {\scriptsize [0.12, 0.24]} \\
Heart Enlarge$\uparrow$ & \textcolor{black!50}{0.00} & \textcolor{black!50}{+0.37} & $-$0.02 {\scriptsize [$-$0.09, 0.04]} & +0.15 {\scriptsize [0.13, 0.16]} & +0.16 {\scriptsize [0.11, 0.22]} & +0.07 {\scriptsize [0.06, 0.08]} & +0.13 {\scriptsize [0.04, 0.22]} \\
Heart Normalise$\uparrow$ & \textcolor{black!50}{0.00} & \textcolor{black!50}{+0.37} & +0.00 {\scriptsize [$-$0.07, 0.07]} & +0.13 {\scriptsize [0.11, 0.16]} & +0.21 {\scriptsize [0.17, 0.25]} & +0.06 {\scriptsize [0.05, 0.07]} & +0.20 {\scriptsize [0.12, 0.28]} \\
\addlinespace[2pt]
\multicolumn{8}{@{}l}{\textit{C2 Containment}} \\
Negated Findings$\downarrow$ & \textcolor{black!50}{--} & \textcolor{black!50}{0.00} & 0.02 {\scriptsize [$-$0.02, 0.05]} & 0.09 {\scriptsize [0.07, 0.11]} & 0.03 {\scriptsize [0.01, 0.04]} & 0.04 {\scriptsize [0.03, 0.06]} & 0.01 {\scriptsize [$-$0.01, 0.03]} \\
Distant Anatomy ($\theta_+$)$\downarrow$ & \textcolor{black!50}{--} & \textcolor{black!50}{0.000} & 2.87 {\scriptsize [2.56, 3.24]} & 0.014 {\scriptsize [0.012, 0.016]} & 0.076 {\scriptsize [0.065, 0.086]} & 0.005 {\scriptsize [0.005, 0.006]} & 1.79 {\scriptsize [1.61, 1.97]} \\
\quad ($\theta_-$)$\downarrow$ & \textcolor{black!50}{--} & \textcolor{black!50}{0.000} & 3.20 {\scriptsize [2.67, 3.85]} & 0.029 {\scriptsize [0.021, 0.038]} & 0.10 {\scriptsize [0.09, 0.13]} & 0.003 {\scriptsize [0.003, 0.004]} & 1.59 {\scriptsize [1.37, 1.82]} \\
\addlinespace[2pt]
\multicolumn{8}{@{}l}{\textit{C3 Reliability}} \\
Seed Agreement ($\kappa_+$)$\uparrow$ & \textcolor{black!50}{0.00} & \textcolor{black!50}{1.00} & 0.70 {\scriptsize [0.63, 0.77]} & 0.70 {\scriptsize [0.64, 0.76]} & 0.10 {\scriptsize [0.00, 0.19]} & 0.36 {\scriptsize [0.27, 0.44]} & 0.62 {\scriptsize [0.46, 0.76]} \\
\addlinespace[2pt]
\multicolumn{8}{@{}l}{\textit{Credibility}} \\
Realism (FD)$\downarrow$ & \textcolor{black!50}{1.910} & \textcolor{black!50}{0.035} & 1.412 & 0.801 & 0.632 & 0.726 & 0.531 \\
Diversity (coverage)$^\dagger\uparrow$ & \textcolor{black!50}{--} & \textcolor{black!50}{0.94} & 0.07 {\scriptsize $\pm$ 0.01} & 0.21 {\scriptsize $\pm$ 0.02} & 0.43 {\scriptsize $\pm$ 0.02} & 0.27 {\scriptsize $\pm$ 0.02} & 0.44 {\scriptsize $\pm$ 0.03} \\
Report Agreement (AUROC)$\uparrow$ & \textcolor{black!50}{0.50} & \textcolor{black!50}{0.85} & 0.52 {\scriptsize [0.51, 0.54]} & 0.66 {\scriptsize [0.65, 0.67]} & 0.74 {\scriptsize [0.73, 0.75]} & 0.62 {\scriptsize [0.61, 0.64]} & 0.73 {\scriptsize [0.72, 0.74]} \\
\quad (drop under Swap)$\uparrow$ & \textcolor{black!50}{0.00} & \textcolor{black!50}{0.35} & 0.04 {\scriptsize [0.02, 0.06]} & 0.17 {\scriptsize [0.15, 0.19]} & 0.24 {\scriptsize [0.22, 0.26]} & 0.13 {\scriptsize [0.12, 0.15]} & 0.22 {\scriptsize [0.21, 0.24]} \\
Report Use ($r_P$, Off-topic) & \textcolor{black!50}{0/$\geq$1} & \textcolor{black!50}{$\approx$0.5} & 1.02 {\scriptsize [1.00, 1.04]} & 0.68 {\scriptsize [0.66, 0.70]} & 0.93 {\scriptsize [0.91, 0.95]} & 0.38 {\scriptsize [0.37, 0.40]} & 0.94 {\scriptsize [0.93, 0.96]} \\
\quad ($r_P$, Scramble) & \textcolor{black!50}{0/$\geq$1} & \textcolor{black!50}{$\approx$0.5} & 1.08 {\scriptsize [1.06, 1.10]} & 0.60 {\scriptsize [0.58, 0.62]} & 0.52 {\scriptsize [0.51, 0.54]} & 0.36 {\scriptsize [0.35, 0.38]} & 0.65 {\scriptsize [0.63, 0.66]} \\
Utility (TSTR AUROC)$\uparrow$ & \textcolor{black!50}{0.50} & \textcolor{black!50}{0.78} & 0.54 {\scriptsize [0.53, 0.56]} & 0.66 {\scriptsize [0.65, 0.67]} & 0.69 {\scriptsize [0.68, 0.70]} & 0.66 {\scriptsize [0.64, 0.67]} & 0.67 {\scriptsize [0.66, 0.68]} \\
\bottomrule
\end{tabular}

%% file: assets/proposals/supplement/tables/effusion_window.tex
\begin{tabular}{@{}l lll ll@{}}
\toprule
 & \multicolumn{3}{c}{Within the text window} & \multicolumn{2}{c}{All edits, at 10\,mL} \\\cmidrule(lr){2-4}\cmidrule(l){5-6}
Generator & Effusion Add & Effusion Remove & Seed Agreement ($\kappa_+$) & Effusion Add & Effusion Remove \\
\midrule
GenerateCT & +0.26 {\scriptsize [0.21, 0.32]} & +0.13 {\scriptsize [0.04, 0.23]} & 0.65 {\scriptsize [0.57, 0.73]} & +0.19 {\scriptsize [0.15, 0.23]} & +0.09 {\scriptsize [0.03, 0.16]} \\
MedSyn & +0.57 {\scriptsize [0.52, 0.61]} & +0.41 {\scriptsize [0.34, 0.47]} & 0.70 {\scriptsize [0.64, 0.76]} & +0.56 {\scriptsize [0.51, 0.60]} & +0.43 {\scriptsize [0.37, 0.50]} \\
Text2CT & +0.02 {\scriptsize [0.01, 0.04]} & +0.21 {\scriptsize [0.16, 0.27]} & 0.10 {\scriptsize [0.00, 0.19]} & +0.02 {\scriptsize [0.01, 0.04]} & +0.22 {\scriptsize [0.17, 0.27]} \\
MedSyn2 & +0.47 {\scriptsize [0.43, 0.50]} & +0.27 {\scriptsize [0.23, 0.33]} & 0.36 {\scriptsize [0.27, 0.44]} & +0.47 {\scriptsize [0.43, 0.50]} & +0.30 {\scriptsize [0.24, 0.35]} \\
RAGText2CT & +0.05 {\scriptsize [0.03, 0.07]} & +0.18 {\scriptsize [0.12, 0.24]} & 0.62 {\scriptsize [0.46, 0.76]} & +0.03 {\scriptsize [0.01, 0.05]} & +0.19 {\scriptsize [0.13, 0.26]} \\
\bottomrule
\end{tabular}

%% file: assets/proposals/supplement/tables/rag_theta.tex
\begin{tabular}{@{}lrl@{}}
\toprule
Additions & Cases & $\theta_+$ \\
\midrule
All additions & 396 & 1.79 {\scriptsize [1.61, 1.97]} \\
Another training CT retrieved & 249 & 2.96 {\scriptsize [2.76, 3.17]} \\
Same training CT retrieved & 147 & 0.026 {\scriptsize [0.020, 0.033]} \\
\bottomrule
\end{tabular}

%% file: assets/proposals/supplement/tables/rewording.tex
\begin{tabular}{@{}l lll ll@{}}
\toprule
 & \multicolumn{3}{c}{Effusion} & \multicolumn{2}{c}{Heart} \\\cmidrule(lr){2-4}\cmidrule(l){5-6}
Generator & Presence & Volume & Classifier & Volume & Classifier \\
\midrule
GenerateCT & inconclusive & inconclusive & inconclusive & -- & -- \\
MedSyn & inconclusive & follows & follows & depends & follows \\
Text2CT & -- & inconclusive & -- & inconclusive & inconclusive \\
MedSyn2 & follows & follows & follows & \textbf{follows} & -- \\
RAGText2CT & -- & -- & -- & depends & -- \\
\bottomrule
\end{tabular}

%% file: assets/proposals/supplement/tables/edit_auroc_seed1.tex
\begin{tabular}{@{}l rrl rrl@{}}
\toprule
 & \multicolumn{3}{c}{Macro AUROC, 18 findings} & \multicolumn{3}{c}{Effusion AUROC} \\\cmidrule(lr){2-4}\cmidrule(l){5-7}
Generator & Original & Edited & Difference & Original & Edited & Difference \\
\midrule
GenerateCT & 0.501 & 0.501 & +0.001 {\scriptsize [$-$0.017, 0.018]} & 0.33 & 0.59 & +0.26 {\scriptsize [0.21, 0.31]} \\
MedSyn & 0.614 & 0.609 & $-$0.004 {\scriptsize [$-$0.017, 0.009]} & 0.11 & 0.70 & +0.59 {\scriptsize [0.55, 0.64]} \\
Text2CT & 0.698 & 0.697 & 0.000 {\scriptsize [$-$0.012, 0.012]} & 0.09 & 0.18 & +0.10 {\scriptsize [0.07, 0.13]} \\
MedSyn2 & 0.591 & 0.595 & +0.004 {\scriptsize [$-$0.009, 0.016]} & 0.16 & 0.61 & +0.45 {\scriptsize [0.41, 0.50]} \\
RAGText2CT & 0.687 & 0.689 & +0.001 {\scriptsize [$-$0.011, 0.014]} & 0.08 & 0.20 & +0.12 {\scriptsize [0.08, 0.16]} \\
\bottomrule
\end{tabular}

%% file: appendix/e04_toolkit.tex
\section{Toolkit and Held-Out Generator}
\label{sec:supp-toolkit}

\subsection{Pre-Flight Checks}
\label{sec:supp-checks}

Before the evaluation starts, six checks are performed on the volumes produced by the generator (\cref{tab:checks}).
Their results are reported at the top of the Report Card, where each check is marked as passed (\checkmark) or flagged
(\textbf{!}). When a check is flagged, the card also indicates the evaluation measures that may be compromised.

\begin{table*}[t]
\centering\footnotesize
\caption{\textbf{Pre-Flight Checks on the volumes produced by the generators.} \checkmark\ passed, \textbf{!} flagged. Spacing: median lung width (real CTs: 234--321\,mm); 1$\times$1$\times$1: every CT has this placeholder spacing. HU: difference of the
median of each flagged landmark from that of the 940 real CTs (one CTFlow CT lacks air, lung and soft-tissue
landmarks). Pose: header, the header contradicts the anatomy; flip, axes flipped for the evaluators that read only the voxels. Seed and Text Window: cases with identical CTs (across seeds; with and without the edit) over cases tested, effusion\,$\cdot$\,heart edits. $^\dagger$Held out, Lite
suite.}
\label{tab:checks}
\input{assets/proposals/supplement/tables/checks.tex}
\end{table*}

\begin{figure}[!tb]
\centering
\includegraphics[width=\linewidth]{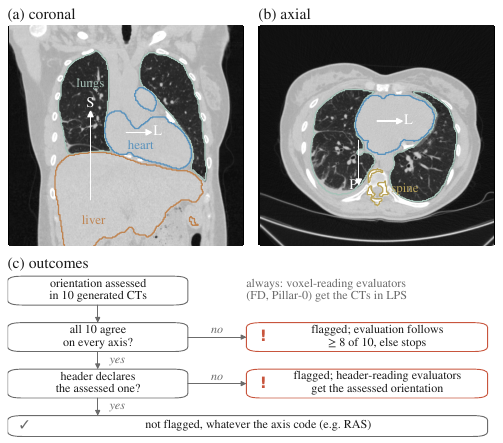}
\caption{\textbf{Pose check.} (a, b) A real CT, patient's left on the image right, with the masks the check reads.
(c) The three outcomes.}
\label{fig:pose}
\end{figure}

\begin{itemize}\setlength{\itemsep}{1pt}\setlength{\parskip}{0pt}
\item \textbf{Complete.} This check verifies that every text--seed pair of the Prompt Suite has a generated CT. If it
is flagged, the whole process is interrupted and the evaluation does not start.
\item \textbf{Spacing.} A generator produces its volumes on the voxel grid it was trained on, but the spacing in
the header of a saved CT is only the one its saving code writes. When the code does not write it explicitly,
SimpleITK and nibabel set a default of 1$\times$1$\times$1\,mm, which may differ from the spacing the generator
actually uses: MedSyn2, trained at 0.7\,mm, and MedSyn save their CTs this way (\cref{tab:checks}). This check
assesses this consistency: it is flagged when every CT has a spacing of 1$\times$1$\times$1\,mm, or when the median
lung width (left--right extent of the lungs, in mm) falls outside the range of real CTs (234--321\,mm,
2.5th--97.5th percentiles). The toolkit resamples every CT to the grid of each evaluator from the spacing in the header, so a flagged spacing must be corrected in the header before measuring, as done here for MedSyn and MedSyn2.
\item \textbf{HU.} This check tests whether the intensities of the generated CTs are calibrated consistently with
real CTs. It flags an intensity landmark (air outside the body; lung, fat and soft-tissue histogram peaks) whose median
differs from that of the 940 real CTs by more than $\max(\mathrm{IQR}, 25\,\mathrm{HU})$, with IQR its interquartile
range across real CTs (air 8, lung 75, fat 15, soft tissue 10\,HU). Any floor between 15 and 35\,HU flags the same
generators, so the outcome does not depend on this choice. If it is flagged, the evaluation proceeds and the toolkit does not rescale the intensities.
\item \textbf{Pose.} This check compares the orientation declared in the header of the generated CTs (axis codes,
\eg LPS, RAS) with the actual orientation of the anatomy in the volume (\cref{fig:pose}). The anatomical orientation
is assessed in 10 generated CTs segmented by
TotalSegmentator~\cite{Wasserthal_2023_TotalSegmentator}: the heart gives the left--right direction, the spine the
antero-posterior one, and the lungs above the liver the cranio-caudal one. This assumes the heart on the left, true
except in situs inversus (about 1 in 10{,}000 people~\cite{Eitler_2022_SitusInversus}). On real CTs, the assessment
returns LPS, the DICOM patient convention. The reports confirm it independently: one-sided effusions effectively lie on the side
the report states. The check has three outcomes (\cref{fig:pose}c):
\begin{itemize}\setlength{\itemsep}{0pt}
\item If the 10 CTs disagree on the direction of an axis, the check is flagged. The evaluation then uses the
direction found in at least 8 of the 10 CTs, and stops if no direction reaches this majority.
\item If the header declares an orientation different from the assessed one, the check is flagged and the toolkit
handles it: the evaluators that read the header (TotalSegmentator, NV-Reason-CT) receive the CTs with the assessed
orientation written in the header. This happens when the saving code writes a fixed orientation that does not
follow the array.
\item If the header agrees with the assessed orientation, the check is not flagged, even when the orientation is
not LPS, since this is only a convention of the generator.
\end{itemize}
In every case, the evaluators that read only the voxels (FD, Pillar-0) receive the CTs reoriented to LPS.
\item \textbf{Seed.} This check detects a generator that ignores the seed, \ie one whose CTs at two seeds are
identical for at least one report (each edit case is generated at three seeds). If it is flagged, Seed Agreement
holds by construction, and Distant Anatomy, which compares the change due to the edit with the variation across
seeds, cannot be computed.
\item \textbf{Text Window.} This check determines whether the edited sentence of each report falls within the text
window of the generator, \ie the number of tokens its text encoder reads. Since text encoders and their tokenizers
differ widely across generators, the check is carried out on the generated CTs: an edit is flagged when the CT of the
edited report is voxel-identical to the CT of the original report at the same seed. Flagged edits count as failures. In the five benchmarked generators, tokenizing the texts shows that these are
exactly the edits beyond the window of the text encoder.
\end{itemize}

\subsection{Report Card}
\label{sec:supp-card}

\begin{figure*}[p]
\centering
\includegraphics[width=0.325\linewidth]{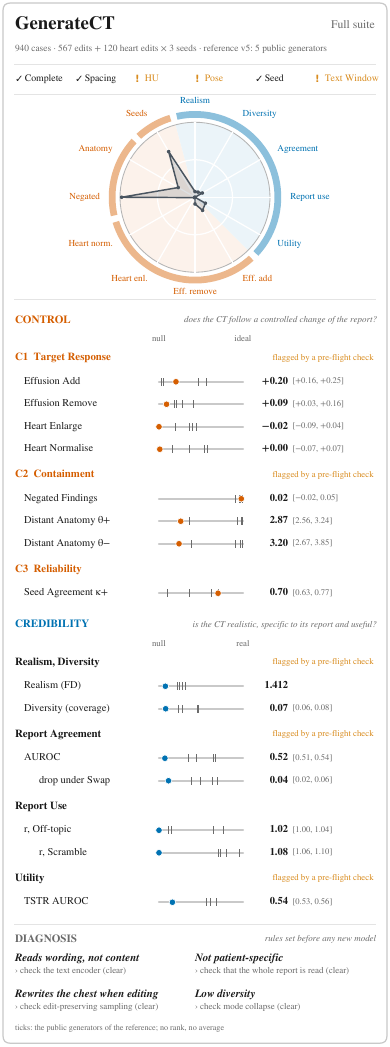}\hfill\includegraphics[width=0.325\linewidth]{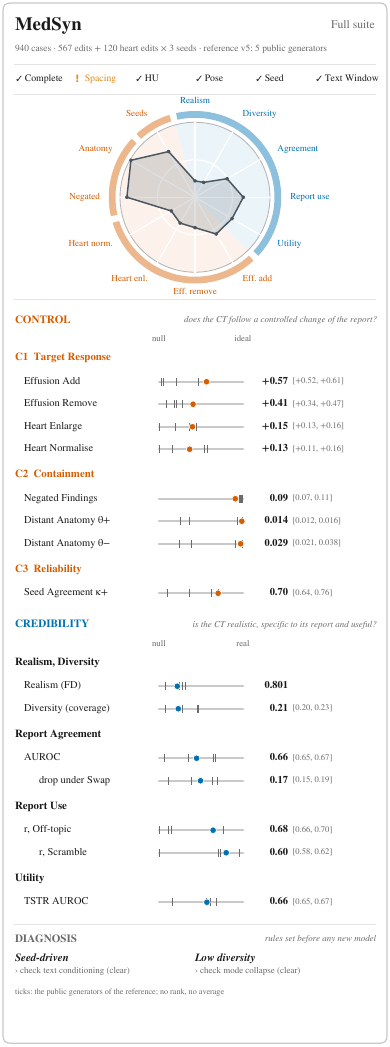}\hfill\includegraphics[width=0.325\linewidth]{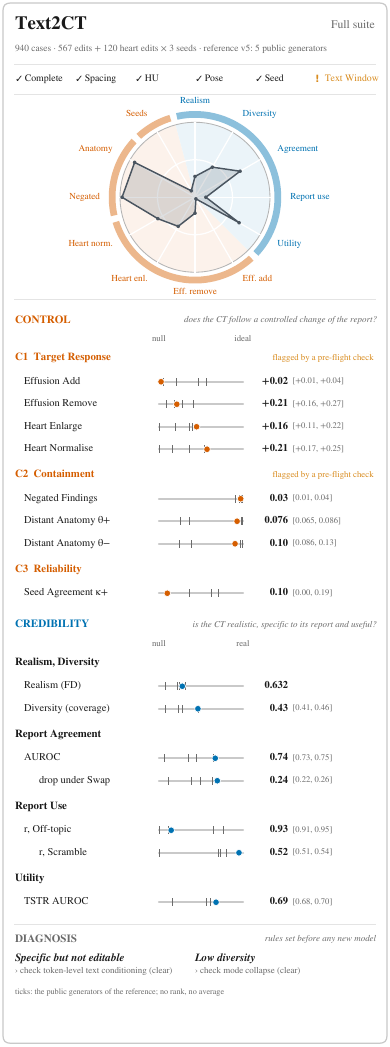}
\caption{\textbf{Report Cards written by the toolkit for the five benchmarked generators} (Full suite; continued in \cref{fig:supp-cards-b}): Pre-Flight Checks, the twelve measures of Tab.~2 as in Fig.~5 with their 95\%
intervals, ticks for the other benchmarked generators, and the triggered rules.}
\label{fig:supp-cards}
\end{figure*}

\begin{figure*}[p]
\centering
\includegraphics[width=0.325\linewidth]{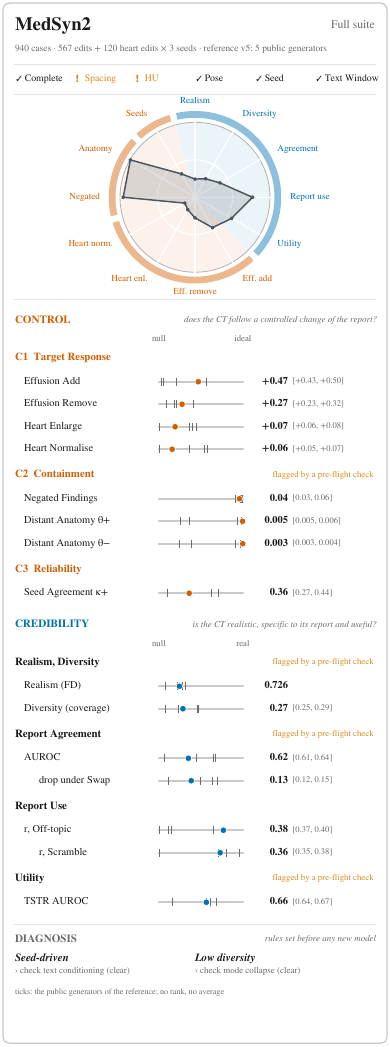}\hfill\includegraphics[width=0.325\linewidth]{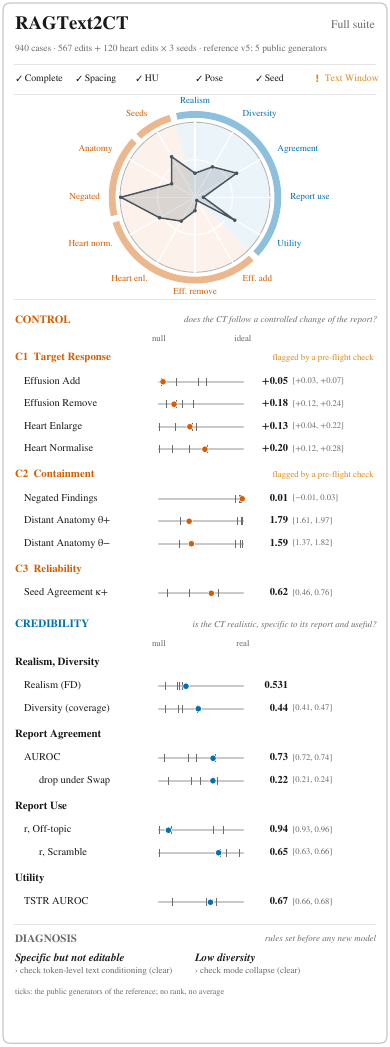}\hfill\includegraphics[width=0.325\linewidth]{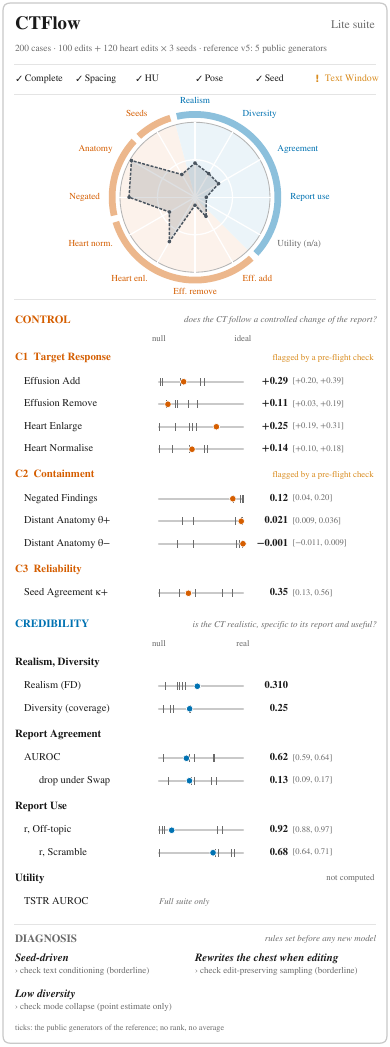}
\caption{\textbf{Report Cards, continued.} MedSyn2 and RAGText2CT on the Full suite; the held-out CTFlow on the Lite
suite, three seeds on the edit cases; Utility is computed on the Full suite only (left out of the profile).}
\label{fig:supp-cards-b}
\end{figure*}

\emph{Measures.} The Report Card (\cref{fig:supp-cards,fig:supp-cards-b}) shows every measure in two ways: as a
bar, in the Control and Credibility panels, and as a spoke of the profile at the top. Since the measures have
different units and ranges (FD and $\theta$ have no upper bound, AUROC lies between 0.5 and 1, the C1 effects
between $-1$ and 1), the card maps each measure $x$ to a position $u\in[0,1]$, with $u=0$ at its null control $x_0$
and $u=1$ at its real or ideal value $x_1$ (grey columns of \cref{tab:supp-main-ci}), clipped to $[0,1]$. These two
anchors are fixed for each measure and are the same for every generator (\cref{tab:card-scale}). Each bar runs from $u=0$ to $u=1$ and shows the position
of the generator with its 95\% confidence interval, where the measure has one (\cref{sec:supp-measures}), next to
the measured value; ticks on the bar mark the positions of the benchmarked generators, from the frozen reference
release. The profile is a radar chart with one spoke for each measure of Fig.~2, from $u=0$ at the centre to $u=1$
at the rim. The variants indented in Tab.~2 ($\theta_-$, $r$ under Scramble) appear as bars only. For Report
Agreement, the spoke shows the drop under Swap rather than the AUROC, because the drop isolates the part of the
agreement that is specific to the CT's own report: it measures how much better a CT agrees with its own report than
with the report of another case, and is 0 when it agrees with any report as well as with its own.

\begin{table}[!t]
\centering\footnotesize\renewcommand{\arraystretch}{1.2}
\caption{\textbf{Positions on the Report Card.} --: no null control. $\bar r$: $r$ clipped to $[0,1]$. Realism is
linear in $\log\mathrm{FD}$, since FDs are compared by ratios; $u=\tfrac12$ at the geometric mean of the anchors, real
CTs blurred by 4\,mm (1.910) and real CTs (0.035). $\theta$ has no upper bound; $u=\tfrac12$ where the shift due to
the edit equals the variation due to the seed, and $u\to0$ as $\theta$ grows. Negated Findings reaches $u=0$ when
every negated finding appears. Report Use fails at both ends (\cref{sec:supp-reading}).}
\label{tab:card-scale}
\begin{tabular}{@{}>{\raggedright\arraybackslash}p{0.255\linewidth}@{\hspace{3pt}}c@{\hspace{3pt}}c@{\hspace{3pt}}c@{\hspace{3pt}}c@{}}
\toprule
Measures & $u$ & \begin{tabular}[c]{@{}c@{}}$u=0$\\(null control)\end{tabular} & $u=\tfrac12$ & \begin{tabular}[c]{@{}c@{}}$u=1$\\(real/ideal)\end{tabular} \\
\midrule
C1, Seed Agreement, Diversity, Report Agreement, Utility & $\frac{x-x_0}{x_1-x_0}$ & $x_0$ & $\tfrac12(x_0+x_1)$
  & $x_1$ \\
Realism & $\frac{\log(x_0/x)}{\log(x_0/x_1)}$ & FD 1.910 & FD 0.26 & FD 0.035 \\
Distant Anatomy & $\frac{1}{1+\theta}$ & -- & $\theta=1$ & $\theta=0$ \\
Negated Findings & $1-\mathrm{NF}$ & -- & $\mathrm{NF}=0.5$ & $\mathrm{NF}=0$ \\
Report Use & $1-2\lvert\bar r-\tfrac12\rvert$ & \begin{tabular}[c]{@{}c@{}}$r=0$\\or $r\ge1$\end{tabular} & \begin{tabular}[c]{@{}c@{}}$r=0.25$\\or $0.75$\end{tabular} & $r=0.5$ \\
\bottomrule
\end{tabular}
\end{table}

\emph{Rules.} Twelve diagnostic rules turn the measures into named failure modes, which the card lists with what to
check (\cref{tab:diagnose}): six describe the behaviour of the generator, five restate the flags of the Pre-Flight
Checks, and one is a note rather than a failure (Different axis convention). Each rule reads only the values of the
generator under evaluation, never those of other generators: the positions $u$ of its measures, its Report Use $r$,
and the report share of the feature variance over cases and seeds (Pillar-0, App.~A). A rule is triggered when its
condition holds for the measured values, and borderline when it holds for the measured values but not for every
value within their 95\% confidence intervals, \ie when the outcome is uncertain. We calibrated the rules, conditions
and thresholds, on the five benchmarked generators and applied them unchanged to the held-out CTFlow; none is
triggered on real CTs (\cref{tab:diagnose}).

\subsection{Lite Suite}
\label{sec:supp-lite}

The Lite suite is the light version of the CtrlCT Prompt Suite, for cheaper tests of new generators. It reduces only
the prompts and keeps the rest of the protocol (three seeds per edit case, measures, anchors and rules): a random
subset of the Full suite with 200 cases, 100 effusion edits balanced between additions and removals, and the same
120 heart edits, \ie 1{,}814 CTs per generator instead of 6{,}017. Utility is computed on the Full suite only. To
check that the Lite suite ranks the generators as the Full suite does, we draw it $D=50$ times. With $\tau$
Kendall's rank correlation between the rankings of the five generators on a measure,
\begin{equation}
\bar\tau=\frac1D\sum_{d=1}^{D}\tau\bigl(\mathrm{rank}_{\mathrm{Lite}_d},\mathrm{rank}_{\mathrm{Full}}\bigr)
\end{equation}
ranges from 0.78 to 1.00 across the measures of \cref{tab:lite}.

\begin{table}[!htb]
\centering\footnotesize
\caption{\textbf{Lite against Full}, over 50 Lite draws: mean $\bar\tau$, and share of draws with the same first
generator as Full.}
\label{tab:lite}
\input{assets/proposals/supplement/tables/lite.tex}
\end{table}

\subsection{CTFlow}
\label{sec:supp-ctflow}

CTFlow runs on the Lite suite (\cref{tab:external}) with its authors' inference code and quality check, which redraws
a CT with fewer than 100 slices. \cref{tab:external} averages C1 over the three seeds; Sec.~5 reports it at the
first seed (Effusion Add $+0.24$, and $+0.44$ on the edits it reads). Four rules are triggered, and two of them are
borderline (\cref{tab:diagnose}):
\begin{itemize}\setlength{\itemsep}{1pt}\setlength{\parskip}{0pt}
\item \textbf{Edits not read}: 52 of the 100 edits fall beyond its 128-token window and give the same CT as the
unedited report;
\item \textbf{Low diversity}: recall is at $u=0.12$, below 0.25 (coverage 0.25, at $u=0.36$);
\item \textbf{Seed-driven}, borderline: the report share is 0.13, and $r$ under a new seed (0.93) exceeds $r$ under
Off-topic text (0.92), \ie a new seed changes the CT as much as an unrelated text. The seed also decides which draws
collapse to a single slice and are redrawn: when one text of a case collapses, the other texts at the same seed
collapse too in 75\% of cases;
\item \textbf{Rewrites the chest when editing}, borderline: SSIM beyond 50\,mm is 0.37 after an edit and 0.18
between CTs of different patients ($u=0.23$), \ie the distant anatomy changes almost as much as for another patient.
\end{itemize}

\begin{table*}[tp]
\centering\footnotesize\setlength{\tabcolsep}{3pt}
\caption{\textbf{The twelve diagnostic rules}, fixed before CTFlow was run. $u$: position on the Report Card
(\cref{tab:card-scale}). \checkmark\ triggered; (\checkmark) borderline; -- not triggered; n/a not applicable (no
edits or seeds). Different axis convention is a note, not a failure: the toolkit aligns the axes. $^\dagger$Held out, Lite suite.}
\label{tab:diagnose}
\input{assets/proposals/supplement/tables/diagnose.tex}
\end{table*}

\begin{table*}[tp]
\centering\footnotesize\setlength{\tabcolsep}{4.5pt}
\caption{\textbf{CTFlow against the reference} on the Lite suite (200 cases against the 692 real CTs of other
patients, 100 balanced edits, three seeds on the edit cases), in the format of Tab.~2. Heart rows: the five generators
on the same 120 heart edits of the Full suite. Bottom: the quantities read by the rules of \cref{tab:diagnose}, for all six generators. Bold and
underlined: best and second among the five benchmarked generators.}
\label{tab:external}
\input{assets/proposals/supplement/tables/ctflow.tex}
\end{table*}

\subsection{Release}
\label{sec:supp-release}
CT-RATE (CC BY-NC-SA 4.0) cannot be redistributed, in whole or in part, but data derived from it can be shared within
its privacy terms. An edited report is part of the dataset; synthetic CTs, features and readouts are derived data. We
therefore release the edits instead of the edited texts, with a script that rebuilds the texts from the user's copy of
CT-RATE. We release all data under CC BY-NC-SA 4.0 with the access conditions of CT-RATE, which keeps the
non-commercial terms of MedSyn (CC BY-NC 4.0) and of the MedSyn2 weights, and the code under Apache 2.0 (datasheet in \cref{tab:datasheet}). The release contains 30{,}525 CTs, 6{,}105 per generator: the 6{,}017 of the Full suite (\cref{sec:supp-lite}), the 80 rewordings (\cref{sec:supp-reword}) and 8 CTs of the original reports, at the second and third seeds, of four of the five excluded effusion cases (\cref{tab:supp-sets}); those of the fifth, a heart case, are part of the Full suite.

\begin{table*}[tp]
\centering\footnotesize
\caption{\textbf{Datasheet of the \counterct{} release}~\cite{Gebru_2021_Datasheets}.}
\label{tab:datasheet}
\begin{tabular}{@{}p{0.13\linewidth}p{0.83\linewidth}@{}}
\toprule
Motivation & Measuring the anatomical control of Text-to-CT generators (Sec.~1). \\
Composition & 940 CT-RATE validation cases (782 patients); 567 effusion edits (396 additions, 171 removals), 120 heart
edits (50 enlargements, 70 normalisations), 80 rewordings (79 analysed, \cref{sec:supp-reword}); seeds and text
hashes. 30{,}525 synthetic CTs of five generators (NIfTI, HU) with manifests and readouts. Real-CT reference:
RadImageNet and Pillar-0 features of the 940 real CTs. Reference results of the five generators (Tab.~2, Report
Cards), versioned with the toolkit. \\
Not included & CT-RATE data and the edited texts (rebuilt by a script from the user's copy); CTFlow volumes (no
stated terms). \\
Collection & Released code, weights and settings of each generator (\cref{sec:supp-generators}); seeds and edits fixed
before generation. \\
Preprocessing & None, except the header orientation (\cref{sec:supp-checks}). \\
Uses & Benchmarking. Not for clinical use; MedSyn2 CTs not for training other models. \\
Privacy & No new acquisitions; RAGText2CT CTs follow the anatomy of a training patient; the CT-RATE re-identification
ban applies. \\
Distribution & Data: CC BY-NC-SA 4.0; code: Apache 2.0. \\
Maintenance & Versioned releases; results never changed after release. \\
\bottomrule
\end{tabular}
\end{table*}

%% file: assets/proposals/supplement/tables/checks.tex
\begin{tabular}{@{}lllllll@{}}
\toprule
Source & Complete & Spacing & HU & Pose & Seed & Text Window \\
 & & {\scriptsize lung width} & & & {\scriptsize effusion\,$\cdot$\,heart} & {\scriptsize effusion\,$\cdot$\,heart} \\
\midrule
Real CTs & \multicolumn{6}{l}{reference: 940 CTs; lung width (234--321\,mm) and directions (LPS) on 40} \\
\midrule
GenerateCT & \checkmark{} & \checkmark{} 288\,mm & \textbf{!} air +102, lung +215, fat +70 & \textbf{!} header & \checkmark{} 0/567\,$\cdot$\,0/120 & \textbf{!} 148/567\,$\cdot$\,6/120 \\
MedSyn & \checkmark{} & \textbf{!} 1$\times$1$\times$1, 222\,mm & \checkmark{} & \checkmark{} & \checkmark{} 0/567\,$\cdot$\,0/120 & \checkmark{} 0/567\,$\cdot$\,0/120 \\
Text2CT & \checkmark{} & \checkmark{} 298\,mm & \checkmark{} & \checkmark{} (flip [0, 1]) & \checkmark{} 0/567\,$\cdot$\,0/120 & \textbf{!} 1/567\,$\cdot$\,1/120 \\
MedSyn2 & \checkmark{} & \textbf{!} 1$\times$1$\times$1, 393\,mm & \textbf{!} air +36 & \checkmark{} (flip [1]) & \checkmark{} 0/567\,$\cdot$\,0/120 & \checkmark{} 0/567\,$\cdot$\,0/120 \\
RAGText2CT & \checkmark{} & \checkmark{} 289\,mm & \checkmark{} & \checkmark{} (flip [0, 1]) & \checkmark{} 0/567\,$\cdot$\,0/120 & \textbf{!} 1/567\,$\cdot$\,1/120 \\
CTFlow$^\dagger$ & \checkmark{} & \checkmark{} 279\,mm & \checkmark{} & \checkmark{} & \checkmark{} 0/100\,$\cdot$\,0/120 & \textbf{!} 52/100\,$\cdot$\,13/120 \\
\bottomrule
\end{tabular}

%% file: assets/proposals/supplement/tables/lite.tex
\begin{tabular}{@{}lcc@{}}
\toprule
Measure & $\bar\tau$ & Same first \\
\midrule
Realism (FD) & 1.00 & 100\% \\
Realism (precision) & 0.96 & 80\% \\
Diversity (recall) & 0.78 & 72\% \\
Diversity (coverage) & 0.94 & 72\% \\
Report Agreement (AUROC) & 0.97 & 86\% \\
\quad (drop under Swap) & 0.98 & 94\% \\
Report share (ICC) & 0.97 & 84\% \\
Effusion Add, first seed & 0.91 & 90\% \\
Effusion Remove, first seed & 0.79 & 100\% \\
SSIM beyond 50\,mm & 1.00 & 100\% \\
\bottomrule
\end{tabular}

%% file: assets/proposals/supplement/tables/diagnose.tex
\begin{tabularx}{\linewidth}{@{}l>{\raggedright\arraybackslash}Xcccccc@{\hspace{6pt}}c@{}}
\toprule
Rule & Condition & Real & GenerateCT & MedSyn & Text2CT & MedSyn2 & RAGText2CT & CTFlow$^\dagger$ \\
\midrule
\multicolumn{9}{@{}l}{\textit{Behaviour}} \\
Specific but not editable & drop under Swap $u\ge0.5$ and Effusion Add (first seed) $u<0.25$ & n/a & -- & -- & \checkmark & -- & \checkmark & -- \\
Seed-driven & report share $<0.5$ and $r$ (new seed) $>$ $r$ (Off-topic) & n/a & -- & \checkmark & -- & \checkmark & -- & (\checkmark) \\
Reads wording, not content & $r$ (Scramble) $\ge$ $r$ (Off-topic) $\ge0.75$ & n/a & \checkmark & -- & -- & -- & -- & -- \\
Not patient-specific & drop under Swap $u<0.25$ & -- & \checkmark & -- & -- & -- & -- & -- \\
Rewrites the chest when editing & SSIM beyond 50\,mm $u<0.25$ (0: another patient; 1: unchanged) & n/a & \checkmark & -- & -- & -- & -- & (\checkmark) \\
Low diversity & coverage or recall (Pillar-0) $u<0.25$ & -- & \checkmark & \checkmark & \checkmark & \checkmark & \checkmark & \checkmark \\
\addlinespace[2pt]
\multicolumn{9}{@{}l}{\textit{Pre-Flight Checks}} \\
Edits not read & Text Window flags $\ge5\%$ of the edits & n/a & \checkmark & -- & -- & -- & -- & \checkmark \\
HU calibration off & HU flag & -- & \checkmark & -- & -- & \checkmark & -- & -- \\
Placeholder voxel spacing & Spacing flag & -- & -- & \checkmark & -- & \checkmark & -- & -- \\
Header contradicts the anatomy & Pose flag on the header & -- & \checkmark & -- & -- & -- & -- & -- \\
Seed ignored & Seed flag & n/a & -- & -- & -- & -- & -- & -- \\
\addlinespace[2pt]
\multicolumn{9}{@{}l}{\textit{Note}} \\
Different axis convention & header true, axes other than LPS & -- & -- & -- & \checkmark & \checkmark & \checkmark & -- \\
\bottomrule
\end{tabularx}

%% file: assets/proposals/supplement/tables/ctflow.tex
\begin{tabular}{@{}lccccccc|c@{}}
\toprule
Measure & \textit{Null} & \textit{Real or ideal} & GenerateCT & MedSyn & Text2CT & MedSyn2 & RAGText2CT & CTFlow \\
\midrule
\multicolumn{9}{@{}l}{\textit{C1 Target Response}} \\
Effusion Add$\uparrow$ & 0.00 & 1.00 & +0.25 & \textbf{+0.54} & +0.05 & \underline{+0.49} & +0.02 & +0.29 \\
Effusion Remove$\uparrow$ & 0.00 & 1.00 & +0.09 & \textbf{+0.45} & +0.19 & \underline{+0.35} & +0.22 & +0.11 \\
Heart Enlarge$\uparrow$ & 0.00 & +0.37 & $-$0.02 & \underline{+0.15} & \textbf{+0.16} & +0.07 & +0.13 & +0.25 \\
Heart Normalise$\uparrow$ & 0.00 & +0.37 & +0.00 & +0.13 & \textbf{+0.21} & +0.06 & \underline{+0.20} & +0.14 \\
\multicolumn{9}{@{}l}{\textit{C2 Containment}} \\
Negated Findings$\downarrow$ & -- & 0.00 & \textbf{$-$0.02} & 0.11 & 0.03 & 0.03 & \underline{0.01} & 0.12 \\
Distant Anatomy ($\theta_+$)$\downarrow$ & -- & 0.000 & 2.62 & \underline{0.016} & 0.094 & \textbf{0.005} & 1.46 & 0.021 \\
\quad ($\theta_-$)$\downarrow$ & -- & 0.000 & 3.47 & \underline{0.041} & 0.085 & \textbf{0.005} & 1.42 & $-$0.001 \\
\multicolumn{9}{@{}l}{\textit{C3 Reliability}} \\
Seed Agreement ($\kappa_+$)$\uparrow$ & 0.00 & 1.00 & \textbf{0.88} & \underline{0.75} & 0.24 & 0.43 & $-$0.02 & 0.35 \\
\multicolumn{9}{@{}l}{\textit{Credibility}} \\
Realism (FD)$\downarrow$ & 1.910 & 0.035 & 1.412 & 0.801 & \underline{0.632} & 0.726 & \textbf{0.531} & 0.310 \\
Diversity (coverage)$\uparrow$ & -- & 0.69 & 0.04 & 0.10 & \underline{0.25} & 0.12 & \textbf{0.26} & 0.25 \\
Report Agreement (AUROC)$\uparrow$ & 0.50 & 0.86 & 0.52 & 0.65 & \textbf{0.74} & 0.63 & \textbf{0.74} & 0.62 \\
\quad (drop under Swap)$\uparrow$ & 0.00 & 0.36 & 0.04 & 0.15 & \underline{0.23} & 0.14 & \textbf{0.25} & 0.13 \\
Report Use ($r_P$, Off-topic) & 0/$\geq$1 & $\approx$0.5 & 1.01 & \underline{0.65} & 0.96 & \textbf{0.38} & 0.98 & 0.92 \\
\quad ($r_P$, Scramble) & 0/$\geq$1 & $\approx$0.5 & 1.08 & \underline{0.57} & \textbf{0.55} & 0.36 & 0.66 & 0.68 \\
\multicolumn{9}{@{}l}{\textit{Inputs of the diagnoses (Lite suite)}} \\
Report share (ICC) & -- & -- & 0.64 & 0.31 & 0.51 & 0.13 & 0.63 & 0.13 \\
$r$, new seed & -- & -- & 0.38 & 0.76 & 0.51 & 0.87 & 0.40 & 0.93 \\
SSIM beyond 50\,mm & -- & -- & 0.30 & 0.79 & 0.43 & 0.94 & 0.44 & 0.37 \\
\quad another patient & -- & -- & 0.15 & 0.29 & 0.14 & 0.29 & 0.13 & 0.18 \\
\bottomrule
\end{tabular}